%% file: main.tex
\pdfoutput=1 
\documentclass{article}

\PassOptionsToPackage{numbers, sort&compress}{natbib}

\usepackage[main, final]{neurips_2025}

\usepackage[utf8]{inputenc} 
\usepackage[T1]{fontenc}    
\usepackage{hyperref}       
\usepackage{url}            
\usepackage{booktabs}       
\usepackage{amsfonts}       
\usepackage{nicefrac}       
\usepackage{microtype}      
\usepackage{amsmath}
\usepackage{cleveref}
\usepackage{todonotes}

\usepackage{multirow}
\usepackage{array}
\newcolumntype{C}[1]{>{\centering\arraybackslash}p{#1}}
\usepackage{endnotes}
\usepackage{rotating}
\usepackage{caption, subcaption}
\usepackage{cleveref}
\usepackage{graphicx}
\usepackage{wrapfig}
\usepackage{booktabs}
\usepackage[table,xcdraw]{xcolor}
\usepackage{changepage}
\usepackage{listings}
\usepackage{booktabs}
\usepackage{longtable}
\usepackage{tabularray}
\usepackage{tabularx}
\usepackage{longtable,ltablex}
\usepackage{array}
\usepackage{caption}

\usepackage{tikz}
\usetikzlibrary{arrows.meta, arrows, positioning, shapes.geometric, shapes, fit, backgrounds, calc}
\definecolor{lightgray}{RGB}{245,245,245}
\definecolor{darkgray}{RGB}{90,108,125}
\definecolor{medgray}{RGB}{138,155,168}
\definecolor{textdark}{RGB}{44,62,80}
\definecolor{arrowcolor}{RGB}{52,73,94}

\newcommand{\ECHO}{\textsc{ECHO}}

\title{Zero-Shot Satellite Image Retrieval through Joint Embeddings: Application to Crisis Response}

\author{%
  James Walsh$^{1,2}$\thanks{Equal contribution.}\quad
  William Fawcett$^{1,2}$\footnotemark[1]\quad
  Grace Colverd$^{1,2}$\footnotemark[1]\quad
  Raúl Ramos-Pollán$^{1,3}$\footnotemark[1] \\[0.6em]
  $^{1}$Trillium Technologies \quad
  $^{2}$University of Cambridge \quad
  $^{3}$Universidad de Antioquia \\
}

\begin{document}
\maketitle

\begin{abstract}
Semantic search of Earth observation archives remains challenging. Visual foundation models such as CLAY produce rich embeddings of satellite imagery but lack the natural-language grounding needed for intuitive query, and full contrastive training of a remote-sensing CLIP-style model requires paired data and compute that are unavailable at global scale.
To allow natural-language querying at global scales, we present GeoQuery, a zero-shot retrieval system that sidesteps data and compute constraints through a two-stage semantic and visual search, leveraging a natural-language embedding of a subset (proxy) of global data. Rather than training a joint encoder, we generate language descriptions for a 100k proxy subset of global Sentinel-2 tiles and optimise the description-generation prompt so that distances in the resulting text-embedding space correlate with distances in the frozen CLAY visual-embedding space. Queries are resolved in two stages, with a text-similarity search over the proxy subset followed by a visual nearest-neighbour search over worldwide CLAY embeddings.
On 76 disaster-location queries covering UK floods, US wildfires, and US droughts, GeoQuery achieves 31.6\% accuracy within 50\,km, with the strongest performance on floods (50\% within 50\,km) where terrain features are well captured by RGB embeddings.
Deployed within a crisis-response system called \ECHO{}, GeoQuery identified vulnerable areas during Cyclone Alfred's 2025 approach to Brisbane, with downstream flood simulations reproducing historical patterns. Prompt-aligned proxies offer a practical bridge between EO foundation models and operational retrieval when full contrastive training is out of reach.
\end{abstract}

\section{Introduction}
Earth observation (EO) archives have grown faster than the tooling to query them. Operators searching for disaster-relevant imagery using queries such as ``areas vulnerable to flooding'' and ``recent burn scars'' must today fall back on coordinate-based queries or manual inspection, because the strongest visual foundation models for satellite imagery, such as CLAY~\cite{clay2024foundation}, produce rich embeddings without text grounding.
The natural fix, training a remote-sensing CLIP-style model end-to-end, requires paired image-text data and contrastive compute that are out of reach at global scale.

GeoQuery is a zero-shot retrieval system that achieves natural-language access to global Sentinel-2 imagery without end-to-end contrastive training, by aligning text and image spaces \emph{indirectly} through prompt-aligned proxies. We sample a 100k subset of global tiles, generate a language description for each via a vision-language model, and optimise the description-generation prompt so that pairwise distances in the resulting text embeddings track pairwise distances in the frozen CLAY visual embeddings. A natural-language query is then resolved in two stages. The first performs a text-similarity search over the 100k proxy descriptions, and the second performs a visual nearest-neighbour search over worldwide CLAY embeddings, using the retrieved proxy tiles as visual anchors.

We demonstrate GeoQuery within \ECHO{}, an operational crisis-response system using Agentic Action Graphs (AAGs) to orchestrate complex workflows, where it supported Australia's National Emergency Management Agency during Cyclone Alfred's 2025 approach to Brisbane. Whilst crisis response provides our motivating application, the contribution is the prompt-aligned proxy method itself, together with an empirical map of where indirect alignment suffices and where it does not, such as flood-prone terrain (well captured) and ephemeral wildfire scars in RGB (poorly captured). We situate GeoQuery against three lines of prior work, namely vision-language foundation models, EO-specific representation learning, and LLM-based agentic orchestration.

The remainder of the paper is organised as follows. \Cref{sec:background} reviews vision-language and EO foundation models alongside prior work on agentic orchestration. \Cref{sec:method} describes the construction of the joint embedding space and the two-stage retrieval procedure. \Cref{sec:results} reports the ablation study and the Cyclone Alfred deployment, and \cref{sec:discussion} interprets the failure modes and outlines further work. \Cref{app:geoquery_ablation} gives the full ablation tables, \cref{app:crisis_response,app:tools} document the \ECHO{} crisis-response framework and its tool library, and \cref{app:crisis_sim_studies} presents CrisisSim case studies.

\section{Background\label{sec:background}}
Vision-language models have made image retrieval possible through joint embedding spaces. CLIP~\cite{DBLP:journals/corr/abs-2103-00020} demonstrated cross-modal alignment at scale by training on 400\,M image-text pairs, enabling zero-shot transfer to novel domains, and subsequent work has extended this approach with larger noisy corpora~\cite{jia2021scaling}, captioning-based bootstrapping~\cite{li2022blip}, and improved contrastive objectives~\cite{zhai2023sigmoid}.
Recent EO-specific models build on this foundation but face unique challenges.
CLAY~\cite{clay2024foundation} provides multi-spectral, multi-resolution embeddings through masked autoencoding on Sentinel and Landsat imagery, capturing seasonal and atmospheric variations but lacking text grounding.
Prithvi-EO-2.0~\cite{szwarcman2025prithvieo20versatilemultitemporalfoundation}, trained on 4.2 million global time series samples from NASA's Harmonised Landsat and Sentinel-2 (HLS) archive at 30\,m resolution, performs land-cover classification well but requires task-specific fine-tuning.
SatMAE~\cite{cong2023satmaepretrainingtransformerstemporal} employs temporal and spectral masking strategies tailored to satellite imagery's unique characteristics.
A wider family of EO foundation models, including Scale-MAE~\cite{reed2023scalemae}, SatlasPretrain~\cite{bastani2023satlaspretrain}, SpectralGPT~\cite{hong2024spectralgpt}, and the multi-sensor DOFA~\cite{xiong2024dofa}, addresses spatial scale, dataset coverage, and spectral richness, with a recent survey provided by~\cite{xiao2024foundation} and a unified evaluation suite by GEO-Bench~\cite{lacoste2023geobench}.
GeoRSCLIP~\cite{Zhang_2024} and RemoteCLIP~\cite{liu2024remoteclipvisionlanguagefoundation} adapt CLIP architectures for remote sensing using the 5\,M-pair RS5M corpus and a corpus assembled from heterogeneous box and mask annotations, whilst SkyScript~\cite{wang2024skyscript} and DOFA-CLIP~\cite{xiong2025dofaclip} extend the contrastive paradigm with semantically diverse pairs and multi-sensor inputs, but all rely on paired image-text data that does not exist at a globally consistent scale.

LLM-based orchestration has been applied across scientific domains, building on foundational work in self-taught tool use~\cite{schick2023toolformer}, reasoning-action interleaving~\cite{yao2023react}, and multi-agent conversation frameworks~\cite{wu2024autogen}. ChemCrow~\cite{bran2023chemcrowaugmentinglargelanguagemodels} augments large language models with eighteen chemistry tools to plan and execute syntheses, whilst Coscientist~\cite{boikoAutonomousChemicalResearch2023} drives end-to-end experimental automation, including the optimisation of palladium-catalysed cross-couplings.
ProtAgents~\cite{ghafarollahiProtAgentsProteinDiscovery2024} coordinates protein design workflows across multiple specialised models, and GeoGPT~\cite{ZHANG2024103976} demonstrates applications to geospatial analysis but lacks the human oversight required for high-stakes scenarios.
These systems decompose complex tasks into tool invocations, but crisis response demands additional constraints, such as interpretability, human validation gates, and deterministic verification. 
For this work, we place GeoQuery inside of \ECHO{}, an AAG processor (for more details see \cref{app:agentic_interface}).

Earth observation presents distinctive challenges not encountered in natural image domains. The integration of optical, synthetic-aperture radar (SAR), and hyperspectral modalities, each governed by distinct physical principles and noise characteristics, creates considerable heterogeneity. Temporal variations arising from seasonal cycles, phenological processes, and atmospheric conditions fundamentally alter scene appearance. Moreover, features manifest differently across spatial scales, from sub-metre commercial imagery to 10\,m Sentinel-2 acquisitions, whilst geographic and sensor-specific variations introduce substantial domain shifts. Several benchmarks address specific crisis-relevant tasks at scale, such as flood segmentation in Sen1Floods11~\cite{bonafilia2020sen1floods11}, global flood mapping in WorldFloods~\cite{mateogarcia2021worldfloods}, and post-disaster building damage assessment in xBD~\cite{gupta2019xbd}, but each requires task-specific labels rather than supporting open-ended retrieval.

Our approach addresses these challenges through a two-stage retrieval strategy that applies expensive vision-language model inference to a proxy subset whilst maintaining global coverage via visual embeddings. This enables zero-shot identification of disaster-relevant imagery without requiring extensive paired training data, bridging pre-trained language models with visual representations. Due to computational constraints, we do not perform end-to-end contrastive training. Instead, we optimise text-generation prompts to maximise rank correlation between pairwise distances in pre-computed CLAY visual space and distances in text-embedding space. This indirect alignment is suboptimal but computationally tractable, and motivates the ablations in \cref{sec:results}.

\section{Method\label{sec:method}}
\begin{figure}[t]
    \centering
    \includegraphics[width=.9\linewidth]{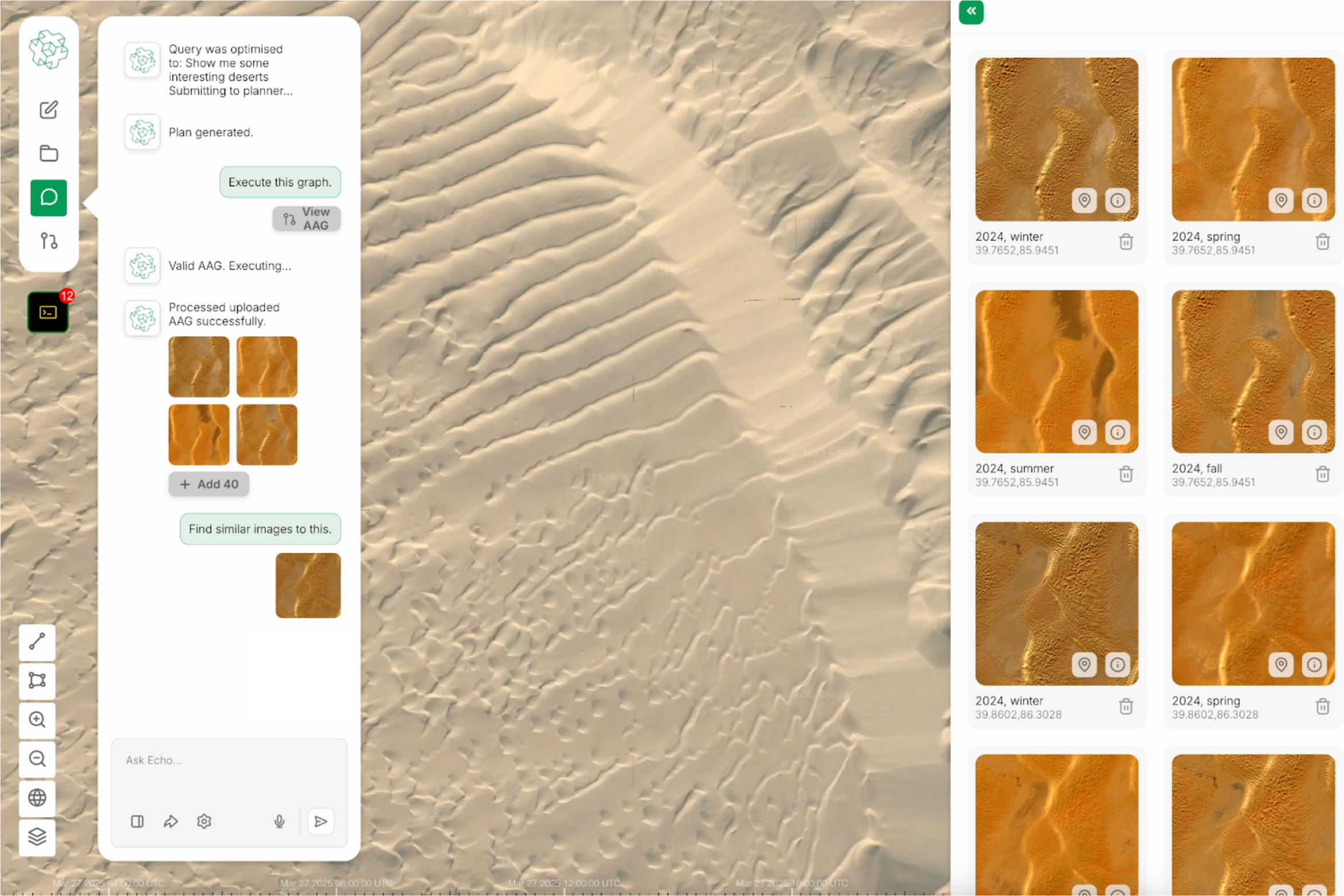}
    \caption{
        The GeoQuery interface within \ECHO, showing the natural-language navigation (``show me deserts'') and the similarity search.
    }
    \label{fig:geoquery_example}
\end{figure}

GeoQuery\footnote{Code: \url{https://github.com/rramosp/geoquery-poc}} allows agents to interact with near-real-time satellite data through natural-language and image search. It provides a natural-language entry point for a satellite image-text foundation model, supporting rapid retrieval of images from the Sentinel-2 satellites as needed. Results are displayed to the user through a custom Cesium~\cite{cesium} integration, with spatial and observational context overlaid on the mosaicked satellite layer, as shown in \cref{fig:geoquery_example}.

The visual embeddings were built by taking 5.12\,$\times$\,5.12\,km image tiles from Sentinel-2 RGB over the median cloud-free pixels per season in 2024 covering the world. Each chip receives four median images, one per three-month period, corresponding to the four seasons (outside the tropics).
These tiles were run through the CLAY encoder~\cite{clay2024foundation} to generate visual embeddings of the full globe. 
To align these image embeddings with natural language, a written representation is required for each image, which is done via a vision-language model and a tailored prompt. 
Since LLMs are expensive to use, we randomly sampled a 100k proxy subset and processed it through a multimodal Gemini~1.5~\cite{geminiteam2024gemini15unlockingmultimodal}, generating text descriptions of what is visible.
We used prompt optimisation~\cite{pryzant2023automaticpromptoptimisationgradient,yang2024opro,khattab2024dspy} to improve the system prompt with which textual descriptions were generated.
We constructed pairs of images and obtained both the text embeddings of their descriptions and CLAY visual embeddings.
The optimisation objective was to achieve a strong rank correlation between distances in the textual and visual embedding spaces, targeting an indirect alignment of the two modalities. GeoQuery summaries reference geological features, possible evidence of past natural disasters, and land-cover information. The 100\,k textual summaries were then embedded into text-embedding space, as for the image embeddings.

\begin{figure}[h!]
\centering
\resizebox{\linewidth}{!}{\input{tikz/embeddings}}
\caption{The structure of GeoQuery's two-level embeddings and search process for the satellite images search.}
\label{fig:geoquery_structure}
\end{figure}
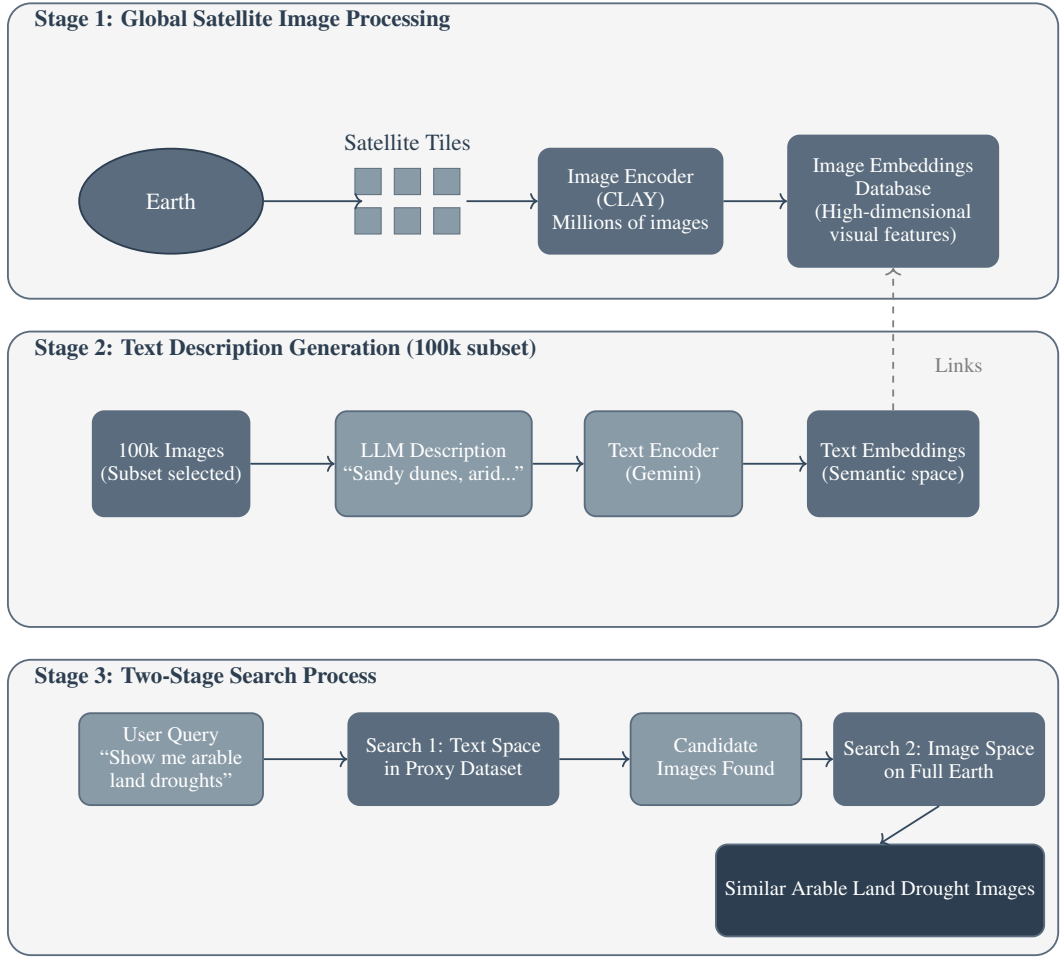

Natural-language queries are then processed in two steps. The first performs a text-similarity search over the proxy subset, and the second uses the retrieved proxy elements as visual queries against the worldwide image embeddings. This cascaded retrieve-then-rerank pattern is well established in dense passage retrieval~\cite{karpukhin2020dpr,khattab2020colbert}. We adapt it here to a cross-modal setting in which the second stage operates over a different embedding space from the first. The arrangement supports cross-modal retrieval at low latency and modest computational cost, and a user may therefore submit a query such as ``areas vulnerable to floods in Brisbane'', without the system having been trained on those specific coordinates. A graphical summary of the embedding and search workflow is shown in \cref{fig:geoquery_structure}.

\section{Results\label{sec:results}}
To evaluate GeoQuery's zero-shot retrieval capabilities, we conducted a systematic ablation study using 76 queries across real 2024 disaster events, comprising 40 UK flood queries, 20 US wildfire queries, and 16 US drought queries. Following the geolocation-retrieval evaluation tradition introduced by PlaNet~\cite{weyand2016planet}, we measure success as the fraction of queries for which the retrieved tile centre lies within a fixed great-circle radius of a confirmed disaster location, reporting accuracy at 50\,km and 100\,km. We tested four configurations that vary the balance between text and image search components, denoted \texttt{balanced\_large} (15 text, 30 image results), \texttt{baseline} (10 text, 20 image), \texttt{text\_focused} (20 text, 10 image), and \texttt{image\_focused} (5 text, 30 image). This design allows us to understand the optimal balance between semantic text matching and visual similarity search in our two-stage retrieval pipeline.

For context, random selection from UK tiles would achieve approximately 2\% accuracy within 50\,km, given the spatial distribution of recorded 2024 flood events.
The \texttt{balanced\_large} configuration achieved best overall performance with 31.6\% of queries successfully identifying disaster locations within 50\,km, a threefold improvement over unconstrained global searches, which achieved only approximately 10\% country-level accuracy. Performance varied substantially by disaster type. UK flood detection was the strongest, with 50\% of queries within 50\,km and 70\% within 100\,km, suggesting that flood-prone terrain features (river valleys, low-lying areas, floodplains) are well captured in the embedding space. US disasters proved more challenging. Drought detection achieved 25\% success within 50\,km, whilst wildfire detection achieved 0\% within 50\,km but reached 40\% within 100\,km. Notable successes included Great Billing floods (6.86\,km accuracy) and Kansas drought areas (8.82\,km), whilst the greater geographic scale and diffuse visual signatures of US disasters presented challenges. 
All configurations maintained search times of approximately one second (0.89--1.05\,s), demonstrating the efficiency of our two-stage architecture even with increased result counts.

\begin{figure}[h!]
    \centering
    \includegraphics[width=1.0\linewidth]{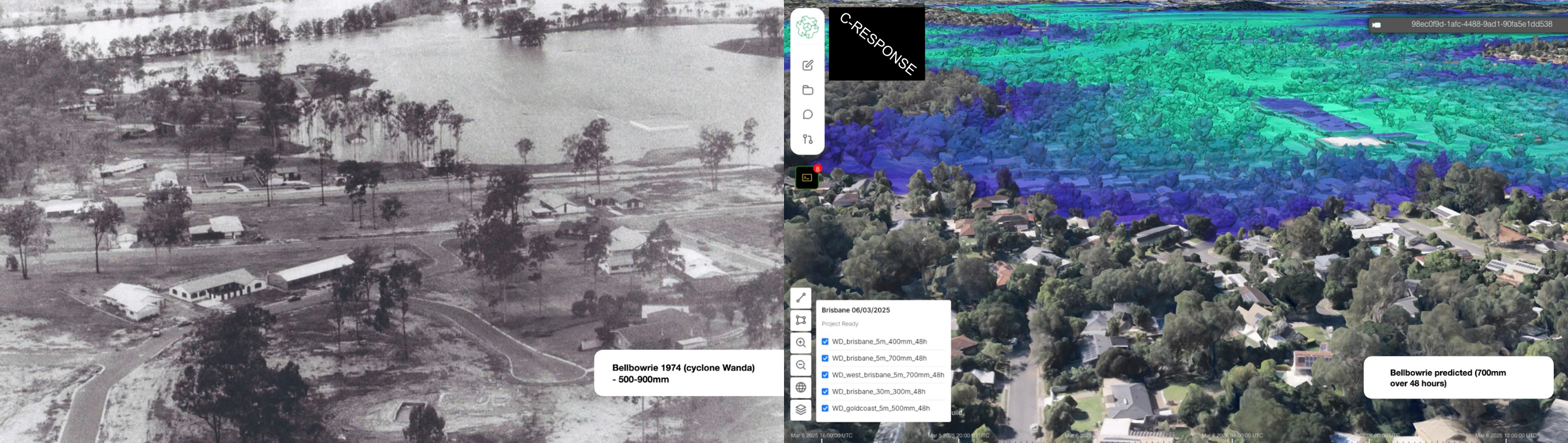}
    \caption{Images of the Bellbowrie suburb of Brisbane, Australia.
    Left: photograph from the 1974 Brisbane floods (Cyclone Wanda). Right: CrisisSim simulation using equivalent rainfall (700\,mm/48\,hr) showing accurate reproduction of historical flood extent.}
    \label{fig:echo_aus}
\end{figure}

We validated the practical utility of GeoQuery within the broader \ECHO{} system during Cyclone Alfred's approach to Brisbane in March 2025, supporting Australia's National Emergency Management Agency. 
Using GeoQuery to identify flood-prone areas and CrisisSim to orchestrate the simulations, we generated flood extent predictions 48 hours before projected landfall.
To assess accuracy, we compared our simulation against the well-documented 1974 Brisbane floods (Cyclone Wanda)~\cite{bom1974brisbane}, where 500--900\,mm of rainfall produced extensive flooding. \Cref{fig:echo_aus} shows this comparison for the Bellbowrie suburb. Our simulation using equivalent rainfall (700\,mm/48\,hr) closely reproduced the observed historical flood extent, generated within minutes without location-specific training or manual adjustment. 
This demonstrates that learned embeddings can identify vulnerable areas, in turn enabling accurate downstream applications such as flood modelling.

The stronger performance on flood detection compared to other disasters suggests certain terrain features are better represented in current visual embeddings, pointing to opportunities for targeted improvement. 
Failure analysis reveals systematic biases. Urban floods are better detected than rural (65\% vs 35\% respectively), suggesting that the model relies on infrastructure cues.
Wildfire detection fails completely at 50\,km, likely because burn scars in RGB imagery are temporally ephemeral and spectrally subtle without near-infrared bands. Detailed performance metrics are provided in \cref{app:geoquery_ablation}.

\section{Discussion and Conclusion\label{sec:discussion}}
This work presents a preliminary investigation into two-stage retrieval for Earth observation archives. Our results suggest that even indirect text-image alignment can provide a useful signal for disaster-relevant retrieval.

Performance varies sharply by disaster type, and the failure modes are informative. Floods are spatially persistent and visually distinctive in RGB. For example, river valleys, low-lying terrain, and urban floodplains are stable features that the embedding can latch onto. Wildfires, by contrast, manifest as burn scars that are temporally ephemeral and spectrally subtle without near-infrared bands. Droughts sit between, large in extent but visually diffuse. This pattern points to extending the visual encoder to multispectral CLAY inputs to capture spectral cues for fire and vegetation stress, performing full contrastive training on the existing 100k proxy as a stepping stone towards end-to-end alignment, and applying temporal stacking to detect change-driven events rather than static terrain.

By combining representation learning with interpretable workflows, our approach seeks to be a step towards bridging the gap between EO foundation models and operational deployment, offering a pathway for more accessible Earth observation analysis.

\bibliography{refs}

\appendix

\section{GeoQuery Ablation Study}
\label{app:geoquery_ablation}

\subsection{Experimental Setup}
We evaluated GeoQuery's disaster location identification capability using 76 queries across three categories: 40 UK flood queries (testing 10 major 2024 flooding locations including Stratford-upon-Avon, Birmingham, and Portsmouth), 20 US wildfire queries (five locations including the Smokehouse Creek Fire in Texas and fires across California, Oregon, and Colorado), and 16 US drought queries (four locations covering drought conditions in the Great Plains and the Southwest).

\subsection{Configuration Details}
The four configurations tested represent different balances between text and image retrieval:

\begin{itemize}
\item \texttt{balanced\_large}: Returns the largest result set, with 15 text matches and 30 image results, optimising for recall whilst maintaining semantic precision
\item \texttt{baseline}: Standard configuration with 10 text matches and 20 image results, representing typical search parameters
\item \texttt{text\_focused}: Prioritises semantic matching with 20 text results but only 10 image results, testing pure language understanding
\item \texttt{image\_focused}: Emphasises visual similarity with 5 text results and 30 image results, testing visual pattern recognition
\end{itemize}

\subsection{Detailed Results}
Performance varied substantially by disaster type and geographic region. UK flood detection performed best, with the closest matches at Great Billing (6.86\,km accuracy), Stratford-upon-Avon (11.48\,km), and Shrewsbury (34.20\,km). The \texttt{balanced\_large} configuration achieved a 50\% success rate for UK floods, suggesting that GeoQuery identifies flood-prone characteristics such as river valleys, low-lying urban areas, and floodplains.

Under the same \texttt{balanced\_large} configuration, US disasters proved more challenging, with drought detection performing better (25\% success within 50\,km) than wildfire detection (0\% within 50\,km, 40\% within 100\,km). Notable successes included Kansas drought areas (8.82\,km accuracy) and Texas Panhandle wildfires (60.08\,km). The \texttt{baseline} configuration recovered only half of the wildfire signal at 100\,km (20\%), with the other configurations weaker still, indicating that the wildfire result is sensitive to the proxy result-set size. The overall lower performance may reflect greater geographic scale diversity and more diffuse visual signatures of drought and wildfire compared to flood-prone terrain.

\subsection{Search Time and Efficiency}
All configurations maintained reasonable search times, ranging from 0.90 to 1.05 seconds per query. The \texttt{balanced\_large} configuration achieved optimal performance without computational penalty, suggesting that the two-stage architecture (text similarity followed by image embedding search) scales efficiently even with increased result counts.

\begin{table}[h]
\centering
\caption{GeoQuery performance by configuration and disaster type.}
\label{tab:geoquery_full_results}
\begin{tabular}{llcccc}
\toprule
Configuration & Disaster Type & Mean Distance (km) & <50\,km (\%) & <100\,km (\%) & Search Time (s) \\
\midrule
\texttt{balanced\_large} & UK Floods & 89.2 & 50.0 & 70.0 & 0.89 \\
\texttt{balanced\_large} & US Droughts & 178.3 & 25.0 & 25.0 & 0.91 \\
\texttt{balanced\_large} & US Wildfires & 201.4 & 0.0 & 40.0 & 0.90 \\
\texttt{baseline} & UK Floods & 98.4 & 30.0 & 60.0 & 1.06 \\
\texttt{baseline} & US Droughts & 189.7 & 25.0 & 25.0 & 1.04 \\
\texttt{baseline} & US Wildfires & 218.1 & 0.0 & 20.0 & 1.05 \\
\texttt{text\_focused} & Overall & 245.4 & 10.5 & 36.8 & 0.95 \\
\texttt{image\_focused} & Overall & 261.2 & 5.3 & 36.8 & 0.90 \\
\bottomrule
\end{tabular}
\end{table}

These results must be interpreted in light of the two-stage search over a proxy dataset, which uses costly vision-language model inference for a small sample of satellite imagery and relies on visual embeddings for global coverage. The results would likely improve if text-description generation were extended to the full global tile set, at a correspondingly higher LLM-inference cost.

These findings support the use of GeoQuery for flood-relevant location retrieval and identify wildfire and drought detection as the priority for further work, whether through richer visual embeddings or specialised training data.

\section{Crisis-Response Framework\label{app:crisis_response}}
Here we summarise the crisis-response framework \ECHO{}, within which GeoQuery is deployed.

\subsection{Agentic Interface\label{app:agentic_interface}}

We consider three operator groups: (1) crisis-response professionals, (2) emergency responders, and (3) the public. Each has different requirements but shares the goal of reducing loss of life. Our approach provides operational transparency and rapid verification for experts whilst remaining accessible to all users. 
Our second group must be informed by the first, alongside patterns that enforce actionable localised deployment whilst accounting for the safety of the responders themselves. Finally, a member of the public should be informed by both of the prior groups and, most importantly, require no expert knowledge to quickly determine the risk associated with a possible action available to them. 

\subsubsection{System Design Principles}

Our approach converts natural-language input from a text or voice chatbot interface into relevant geospatial visualisations, complete with any overlaid simulations, warnings, or other situationally relevant information.
This is done through an intermediate executable graph, which defines the totality of the actions to be undertaken to acquire the information and generate the visualisation requested. We define this as an AAG.
Each graph connects a series of ``tools'', each of which can perform a specific operation (e.g. acquire the most recent satellite image for a specified area).

 Confining operations to this format is motivated by transparency and deterministic verifiability to the expert user. 
 Each AAG must be verified and accepted by a human before execution.
 Once a particular scenario is well constructed and considered to be ``understood'' in the narrative of the agentic pipeline, it may then be made available as a knowledge base for our further constituencies. The order of operations is:

\begin{enumerate}
    \item Risk identification via external monitoring (e.g. meteorological alerts for severe rainfall).
    \item The risk is developed into a ``project'' defined spatially and temporally. These extents define the bounds for digital twinning of infrastructure and topography, a core foundation for downstream simulation and scenario building. For example, a national meteorological agency might flag a possible flood event triggered by 48 hours of intense rainfall in Australia.
    \item Once enough information is collected on a given project, experts may begin to define the nature of the inquiry. \ECHO{} supports requests to specify which real-time data streams must be monitored first, simulate crisis events, and finally define alerting procedures as information is ingested. For example, five-metre digital elevation maps are downloaded alongside current precipitation projections and building data for Brisbane.
    \item These highly granular assets are then accessible to an expert to rapidly define the line of geospatial inquiry and identify risks unknown to the automated system. For example, an expert might request a flood model and an evaluation of which buildings may be suitable for sheltering at-risk individuals in place.
    \item A crisis responder or member of the public may then request hyper-localised information from the contextually aware agent. For example, they might ask which roads are likely to be inaccessible to a particular vehicle, such as an ambulance or a family car, when planning a safe route.
\end{enumerate}

For any of the steps above to be possible, we require a means to construct these AAGs, and the tools must also be similarly available.

\subsection{Building Blocks}

\ECHO's architecture consists of three foundational components that transform natural-language queries into actionable crisis intelligence, namely the system design principles prioritising human oversight and transparency, the Agentic Action Graphs as the computational framework for orchestrating complex workflows, and a tool library providing the primitive operations available to AAGs.
Together, these building blocks create a system that is both powerful enough to handle complex geospatial analysis and constrained enough to ensure safety in high-stakes deployments.

\subsubsection{Agentic Action Graphs}
\begin{figure}
\begin{adjustwidth}{-8em}{-8em} 
    \centering
    \input{tikz/query_processing_pipeline}
    \end{adjustwidth}
    \caption{\centering Query processing workflow incorporating graph planning, execution, and validation with expert review.}
    \label{fig:pipeline}
\end{figure}
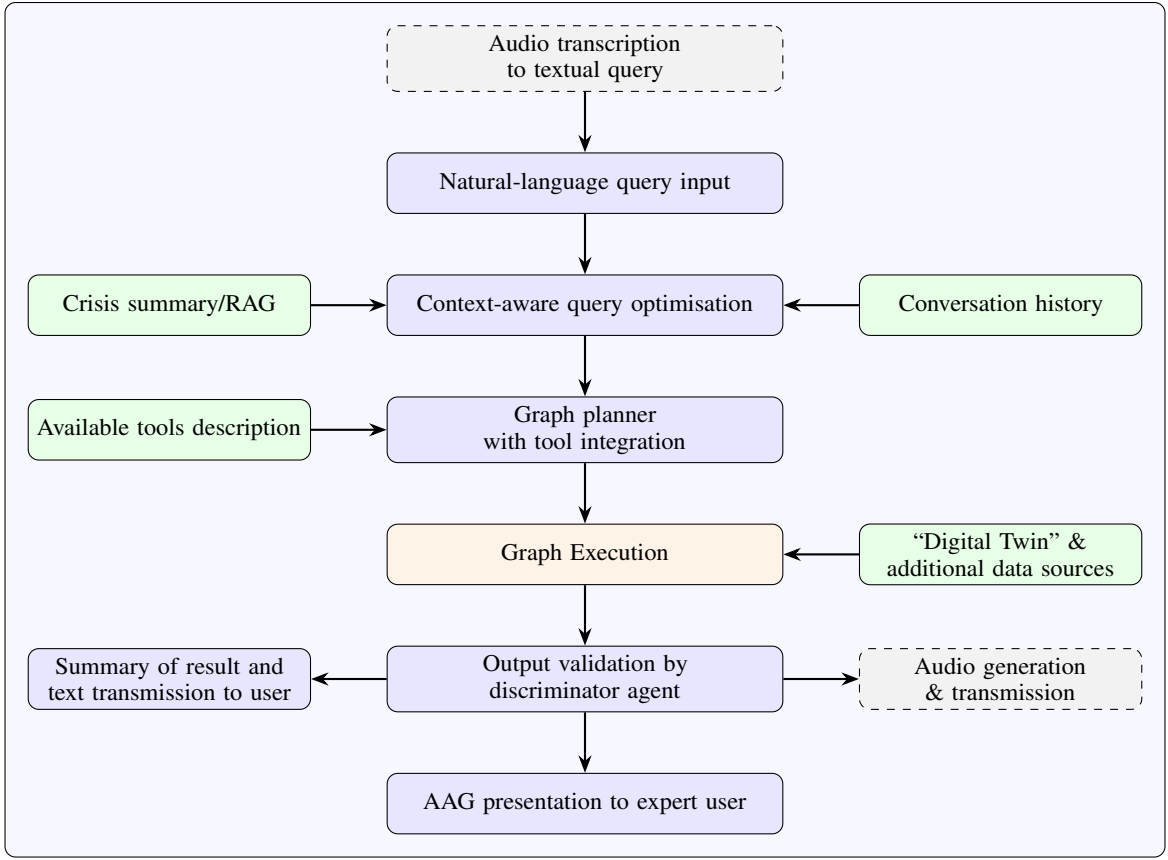

\begin{figure}[h!]
\begin{adjustwidth}{-6em}{-8em} 
\input{tikz/example_AAG}
\end{adjustwidth}
\caption{
    Example AAG for flood modelling. 
    The boxes show the internal tools used by \ECHO. For example, \texttt{DC\_Get\_OSM\_GeoCoding} converts the name `Brisbane' into the relevant geographic data, including a boundary polygon. 
    The AAG is context-aware of the inputs and outputs of each tool, and \ECHO{} can transfer the data from tool to tool. 
    See the tables in \cref{app:tools} for a full list of tools and their descriptions.
}
\label{fig:aag_floods}
\end{figure}
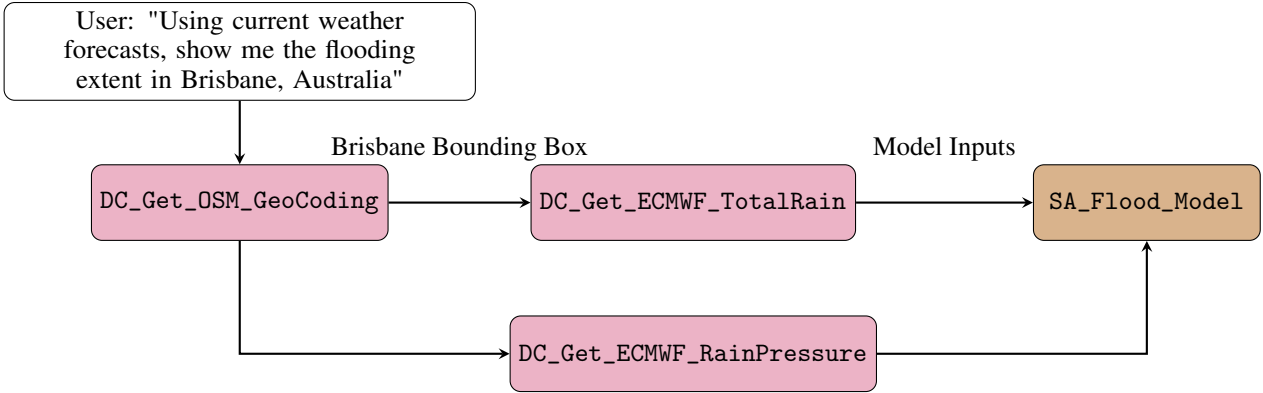

Once a crisis has been localised, data ingestion pipelines can be initiated inside \ECHO{} and additional context uploaded. 
The query processing approach is summarised in \cref{fig:pipeline}. 
An example output for a given flooding query is provided by \cref{fig:aag_floods}.

Formally, we define an AAG as $G = (V, E, \tau, \phi, \sigma)$ where:
$V$ represents vertices (individual tools or operations);
$E \subseteq V \times V$ represents directed edges (data dependencies);
$\tau: V \rightarrow T$ maps vertices to tool types from our tool library $T$;
$\phi: V \rightarrow P$ maps vertices to parameter configurations;
$\sigma: E \rightarrow S$ defines the data schema for edge transitions. The graph must satisfy: (1) acyclicity, (2) type consistency across edges, and (3) human validation checkpoints before the execution of any operation marked as high-risk in $T$.

Currently, these graphs are generated through prompt engineering and minimal RAG, rather than an explicit policy that could be improved via reinforcement learning.
An example of the query optimisation, planner system, and user prompts is provided in \cref{sec:prompt_sys_queryopt,sec:prompt_sys_planner,sec:prompt_usr_planner}.
Most general tools have now been fully defined both in natural language and in their executable code.

With the AAG approach established, we now describe the specific tools that agents can orchestrate within these graphs. Among these tools, two deserve special attention for their novel contributions to crisis response, namely GeoQuery for satellite image retrieval and CrisisSim for flood modelling.

\subsubsection{General Tools\label{sec:model_choice}}
The general tools provide broad geospatial and data collection operations.
This primarily involves geospatial data management and processing, as well as polling designated external APIs.
Many of the geospatial processes are implemented via packages such as \textsc{gdal} \cite{gdal} and \textsc{geopandas} \cite{kelsey_jordahl_2019_3483425}. 
The tools are listed in the tables in  \cref{app:tools}. 
All tools are constructed from an Abstract Base Class, which dynamically generates the input and output formats from the logic provided as the executable function.

\subsubsection{Visualisation Interface}
Finally, \ECHO{} is made accessible via a web interface. 
For the running example of Cyclone Alfred reaching Brisbane, Australia, on 6 March 2025, the resulting flood extent is displayed in \cref{fig:echo_aus}.

Having described \ECHO's architecture and components, we now demonstrate its capabilities through two validation studies, namely a real-world flood prediction scenario and a systematic evaluation of the GeoQuery retrieval system.

\subsection{Query Optimisation System Prompt\label{sec:prompt_sys_queryopt}}
\begin{lstlisting}
    Your job is to understand the user's intent and format their query appropriately. There are two main types of requests:

1. Disaster Risk Analysis
For requests about assessing disaster risks (fire, floods, earthquakes, etc.), ensure the query includes:
- Location of interest
- Time horizon
- Type of disaster

Example 1:
Previous context: Take me to valencia
Current state variables available: {"data": bbox"}  
User Input: Can you determine if this area is flood prone over the next few days? 
Bot: thinking: I have the bbox which defines the current map view
Output: {'state': 'complete', 'query_for_planner': 'Generate the flood risk map for the accompanying bbox for the next 3 days'}

Example 2:
User Input: Can you determine if this area is flood prone? 
Response: {'state': 'gathering_info', 'next_question': 'Which region are you interested in analysing flood risk for and over which time period? '}
User Input: Valencia, over the next week
Output: {'state': 'complete', 'query_for_planner': 'Generate the flood risk map for the accompanying bbox for 100mm of rain'}

Example 3: 
Previous task context: an action has resulted in this action from the Planner Agent: {{
        "type": "ux",
        "instruction": "pan_camera",
        "role": "agent", 
        "data": bbox
    }}
    {{
        "type": "map_layer",
        "instruction": "add_layer",
        "role": "agent", 
        "data": flood_map
    }}

User Input: Can you identify a potential safe zone? 
Bot: Thinking: I have the region and the flood map. I can send this to the planner with instructions to identify a safe public area out of the flood.
Output: {'state': 'complete', 'query_for_planner': 'Find any public POIs like hospitals, firestations, police buildings that are not in the flood risk zone'}

Example 4: 
User Input: Im worried about the floods in my area. Can you identify a safe zone? 
Response: {'state': 'gathering_info', 'next_question': 'Which region are you in. I will default to analysing the risk over the next 12 hours.'}
User Input: I'm in London. Thanks. 
Output: {'state': 'complete', 'query_for_planner': 'Generate the flood risk map for the London over the next 12 hours, and then identify the safe POIs away from the flood.'}

2. Satellite Image Search
For general satellite image queries that don't involve disaster risk (e.g., "Show me images of oceans near deserts"). These queries do not require a time horizon, nor a specific location. Feel confident to pass on such queries to the planner as long as no disasters are mentioned.

Example 1: 
User input:  show me forests
Output: {'state': 'complete', 'query_for_planner': 'Show me forests'}

Example 2:
User input:  I'm interested in volcanoes 
Response: {'state': 'gathering_info', 'next_question': 'Are you interested in satellite images of volcanoes or are you concerned about the short term risk of a volcanic eruption?'}
User Input: 'I want to see images of volcanoes'
Output: {'state': 'complete', 'query_for_planner': 'Show me images of volcanoes'}

Output Format when query needs clarification:
{
    "state": "gathering_info",
    "next_question": "specific question to ask user"
}

Output Format when query is clear:
{
    "state": "complete",
    "query_for_planner": "reformulated user query with all relevant details"
}

Make use of the previous context to construct the query for the planner. 

\end{lstlisting}

\subsection{Planner System Prompt\label{sec:prompt_sys_planner}}
\begin{lstlisting}
You are a computational geography expert. Create a JSON plan showing logical sequence of analysis steps, including parallel steps where possible.

Available Tools:
**{self._format_tools_list()}**

Output Format:
```json
{{
    "reasoning": "string explaining analysis approach",
    "steps": [
        {{
            "id":  {{
                "type": "string",
                "description": "Unique identifier for this step"
            }},
            "tool": {{
                "type": "string",
                "description": "Must match tool name from available list"
            }},
            "purpose": {{
                "type": "string",
                "description": "Brief purpose of this step"
            }},
            "input": {{
                "type": "array",
                "description": "List matching tool's 'in' requirements"
            }},
            "output": {{
                "type": "array",
                "description": "List matching tool's 'out' definition"
            }},
            "after": {{
                "type": "array",
                "description": "IDs of steps this must follow"
            }}
        }}
    ]
}}
```
Rules:
1. Start with OSM_Geocode for location queries
2. Use 'after' for dependencies
3. Empty 'after' means step can start immediately
4. Input/output must match tool definitions exactly
5. Use only listed tools
6. OSM Points of Interest should only be used when looking for specific physical infrastructure tags

**{examples}**

Return only valid JSON matching this format using listed tools.
\end{lstlisting}

\subsection{Planner User Prompt\label{sec:prompt_usr_planner}}
\begin{lstlisting}
Create a logical tool sequence plan for: ```{query}``` 

Here are all previous messages between the user and the planner:
**{conversation_history}**

Here are the previous plans the agent has generated based on previous messages:
**{self.planning_history}**

You should use the conversation history and any previous plans to inform the next plan.

Respond in two phases. First by providing some high level thoughts about the process to follow and then the JSON object. Put the  reasoning inside the xml tags <thinking> and </thinking> and the JSON object inside the xml tags <answer> and </answer>.
\end{lstlisting}




\clearpage
\newpage


\clearpage
\newpage 
\section{Tools\label{app:tools}}
\renewcommand{\tabularxcolumn}[1]{p{#1}}





\input{tables/tools_by_group}

\newpage
\section{CrisisSim Case Studies\label{app:crisis_sim_studies}}

\begin{figure}[h!]
    \centering
    \includegraphics[width=1\linewidth]{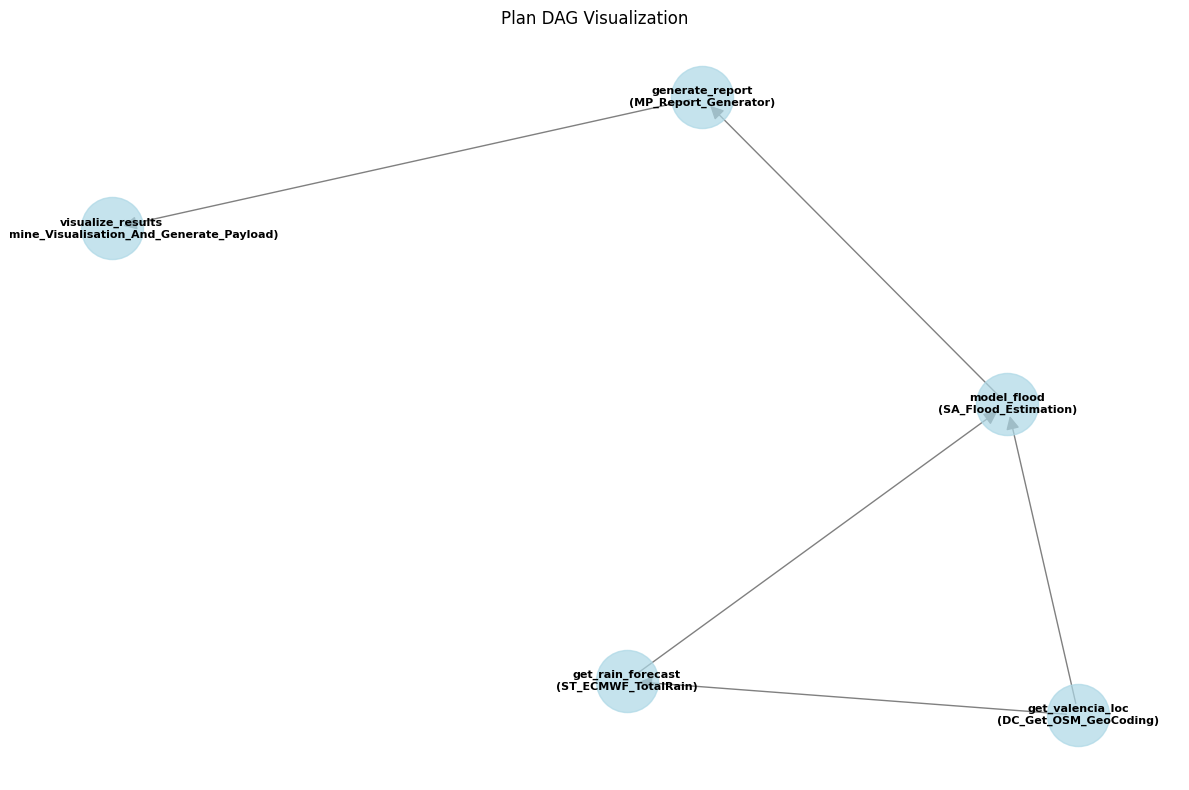}
    \caption{Crisis Centre flood simulation workflow -- Initial disaster preparedness query for Valencia showing the agent's ability to construct comprehensive flood risk assessment plans including meteorological data integration, discharge estimation, and safety zone identification.}
    \label{fig:case_study_1}
\end{figure}

\begin{figure}[h!]
    \centering
    \includegraphics[width=1\linewidth]{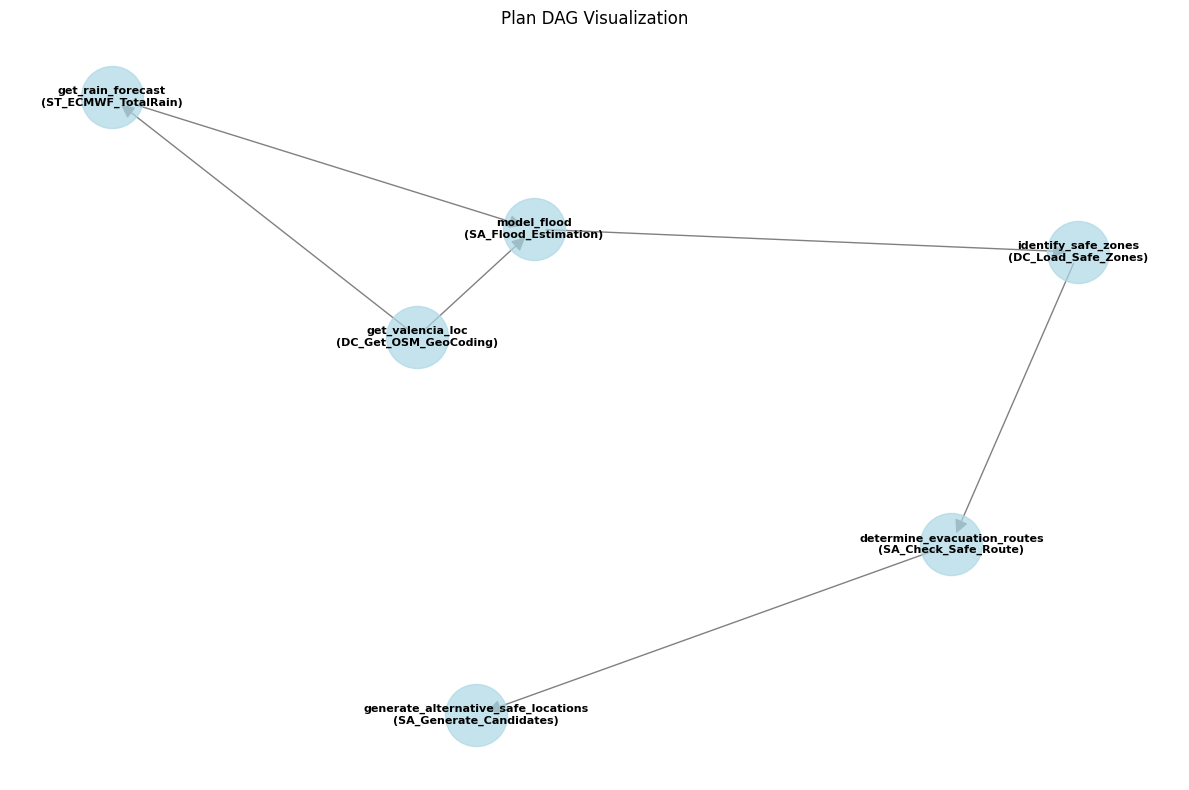}
    \caption{Crisis Centre escalation scenario -- Response to elevated METEO agency alerts demonstrating the system's capability to adapt flood risk assessments based on changing meteorological conditions and generate revised emergency response measures.}
    \label{fig:case_study_2}
\end{figure}

\begin{figure}[h!]
    \centering
    \includegraphics[width=1\linewidth]{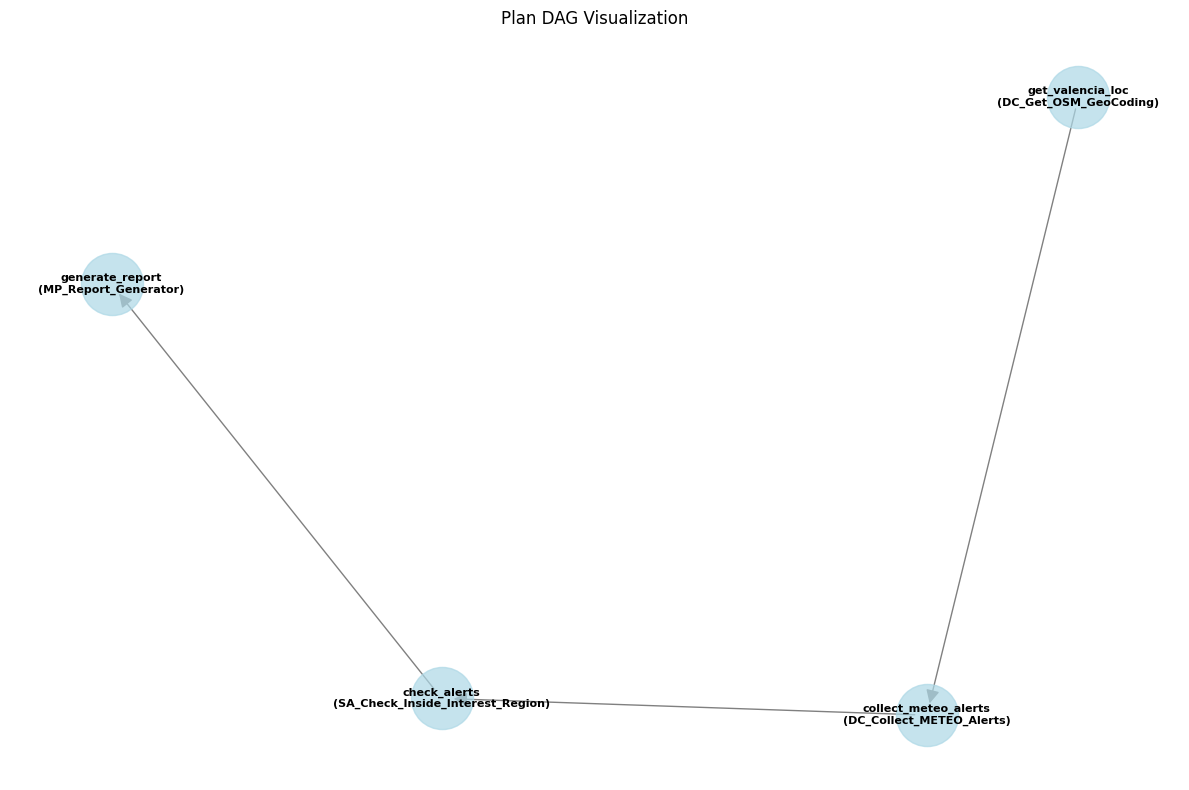}
    \caption{Crisis Centre monitoring and alerting workflow -- Automated collection and assessment of meteorological alerts for Valencia, showing integration with external weather services and real-time risk evaluation capabilities.}
    \label{fig:case_study_3}
\end{figure}

\begin{figure}[h!]
    \centering
    \includegraphics[width=1\linewidth]{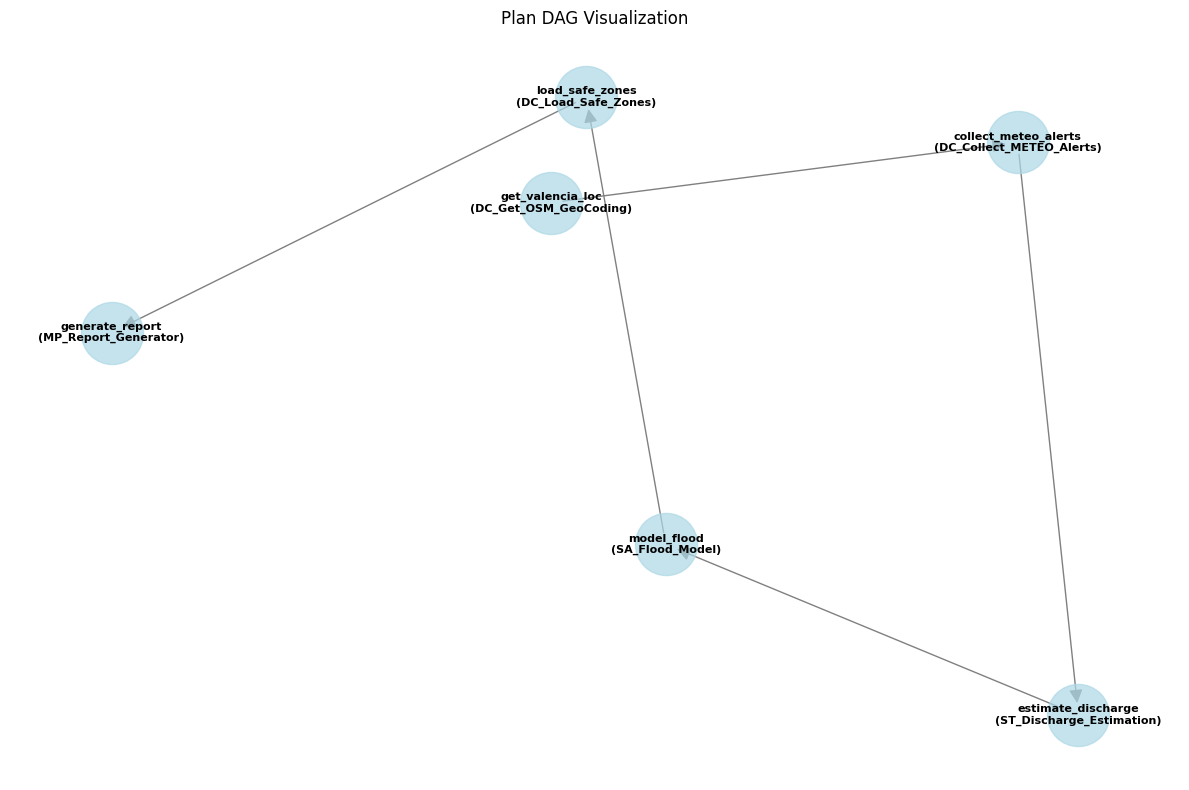}
    \caption{Crisis Centre severe weather response -- Updated flood risk assessment incorporating severe weather alerts with heavy rain predictions, demonstrating the system's ability to revise emergency response measures based on escalating conditions.}
    \label{fig:case_study_4}
\end{figure}

\begin{figure}[h!]
    \centering
    \includegraphics[width=1\linewidth]{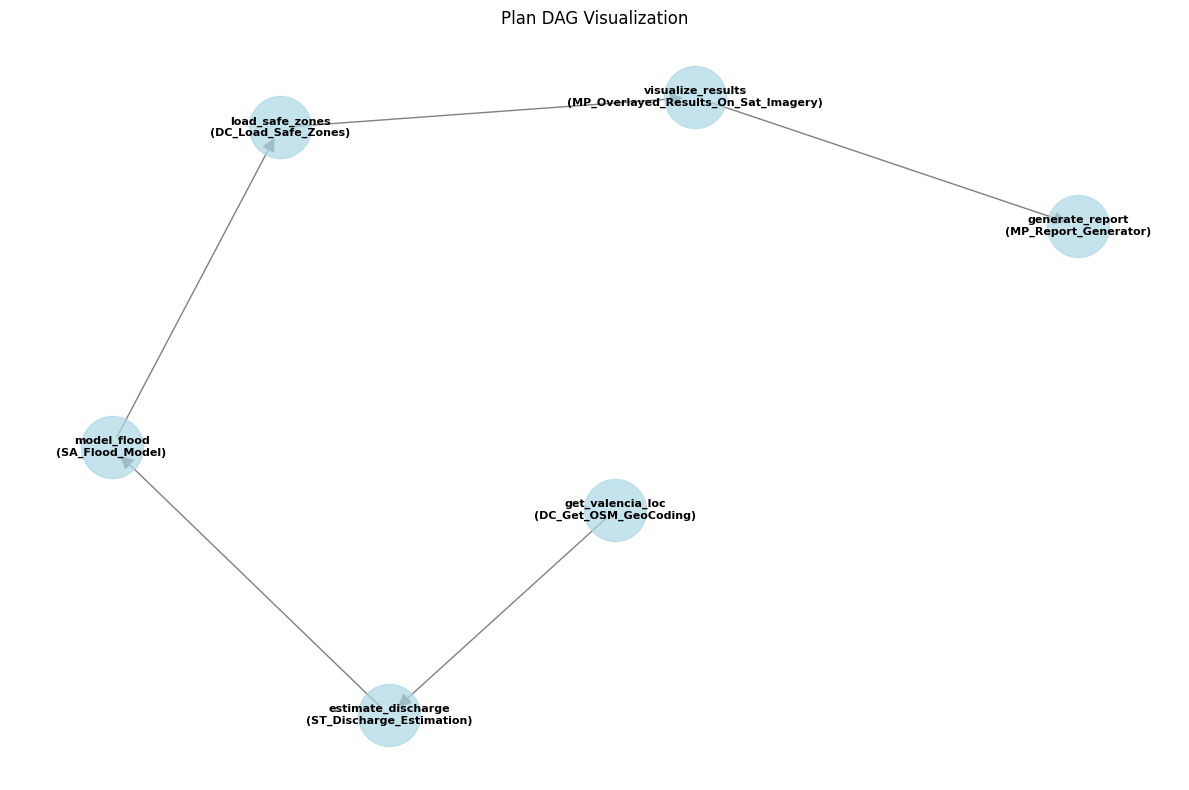}
    \caption{Crisis Centre quantitative flood modelling -- Flash flood simulation based on specific rainfall amounts (200\,mm), showing detailed workflow for discharge estimation, flood extent modelling, safe zone identification, and satellite imagery integration.}
    \label{fig:case_study_5}
\end{figure}

\begin{figure}[h!]
    \centering
    \includegraphics[width=1\linewidth]{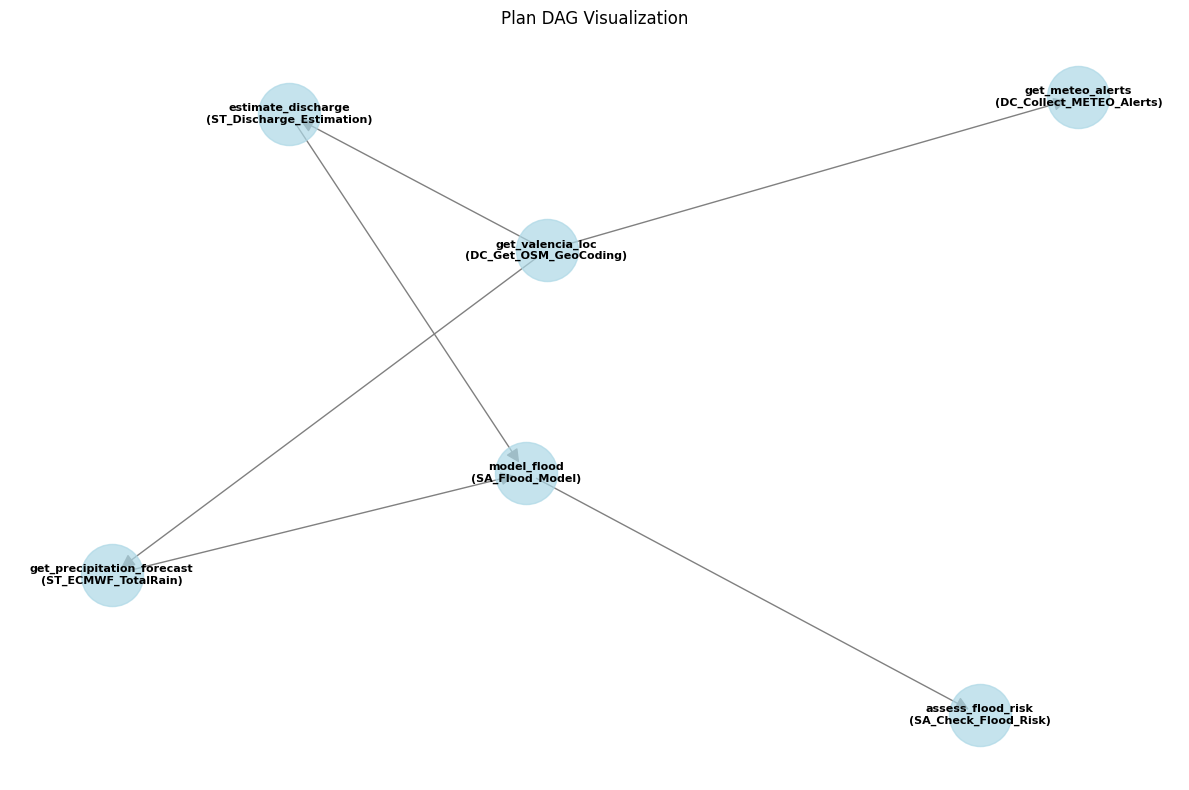}
    \caption{First Responder vehicle safety assessment -- Road network analysis for emergency vehicle navigation during flood conditions, demonstrating integration of meteorological forecasts, discharge modelling, and transportation infrastructure evaluation.}
    \label{fig:case_study_6}
\end{figure}

\begin{figure}[h!]
    \centering
    \includegraphics[width=1\linewidth]{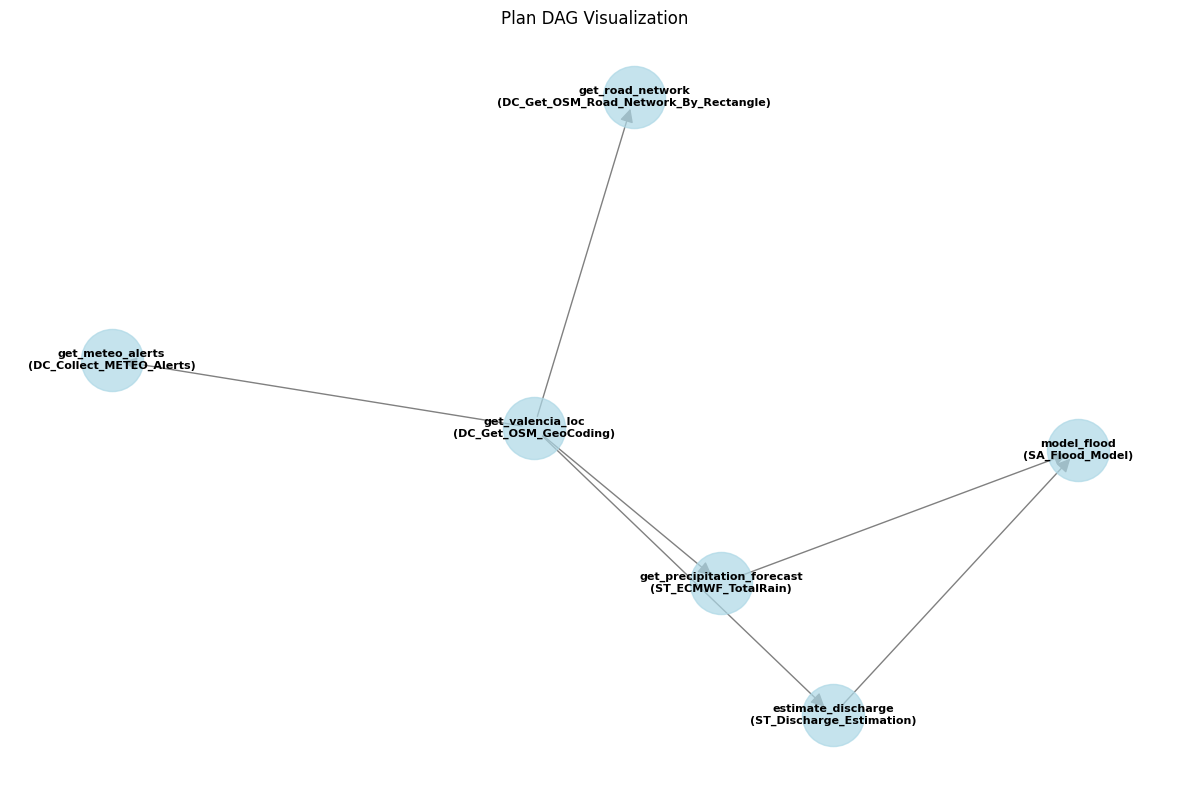}
    \caption{First Responder route planning -- Continuation of vehicle safety assessment showing road network extraction and flood risk evaluation for emergency response vehicle routing during crisis conditions.}
    \label{fig:case_study_7}
\end{figure}

\begin{figure}[h!]
    \centering
    \includegraphics[width=1\linewidth]{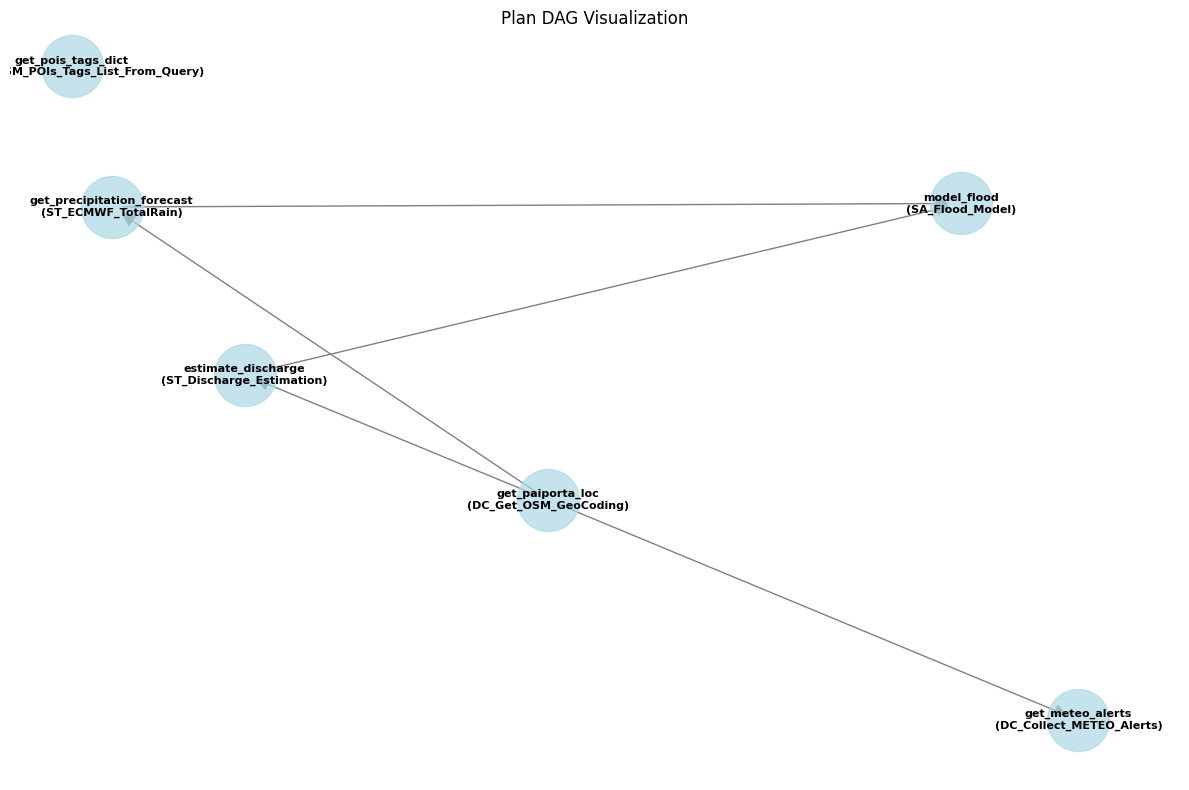}
    \caption{Citizen safe zone identification -- Public-facing workflow for identifying emergency safe zones in Paiporta, showing integration of flood modelling, meteorological alerts, and points-of-interest analysis for civilian evacuation planning.}
    \label{fig:case_study_8}
\end{figure}

\begin{figure}[h!]
    \centering
    \includegraphics[width=1\linewidth]{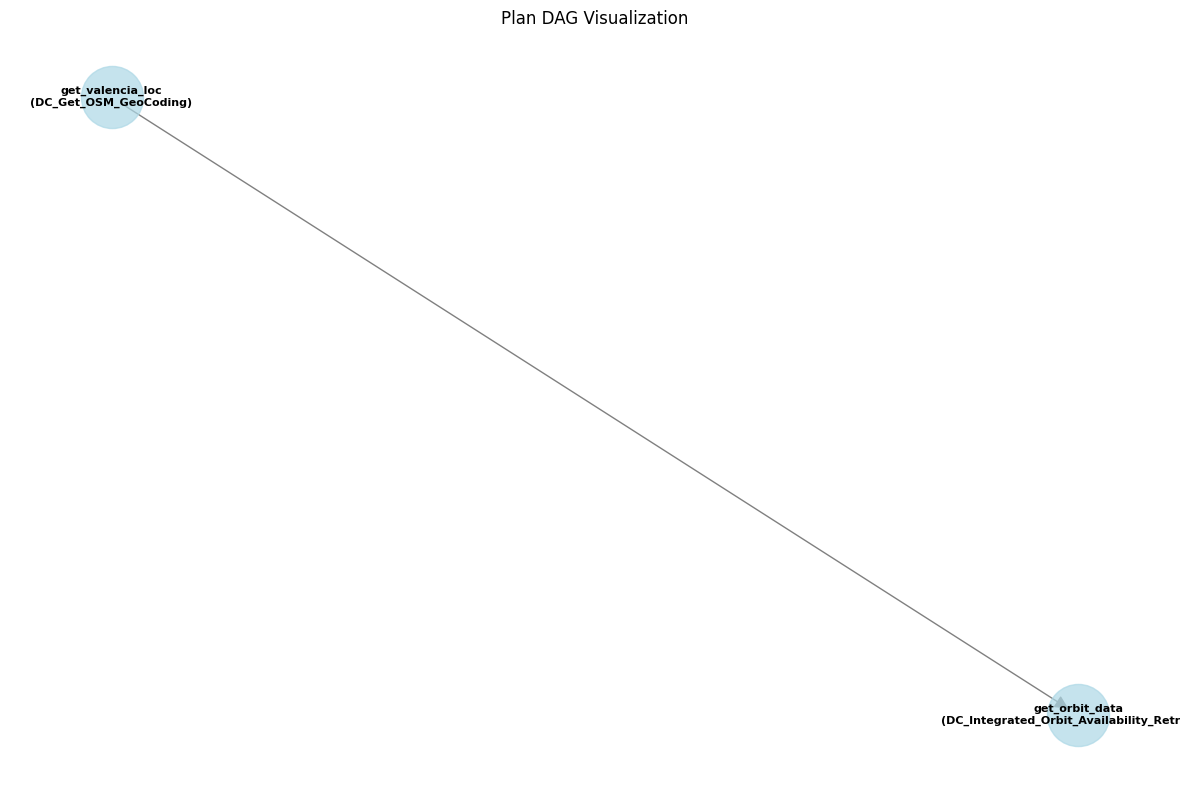}
    \caption{Internal alert reactivity -- Satellite orbit planning and availability assessment for Valencia, demonstrating the system's capability to integrate with satellite mission planning for post-disaster imagery acquisition and analysis.}
    \label{fig:case_study_9}
\end{figure}

\begin{figure}[h!]
    \centering
    \includegraphics[width=1\linewidth]{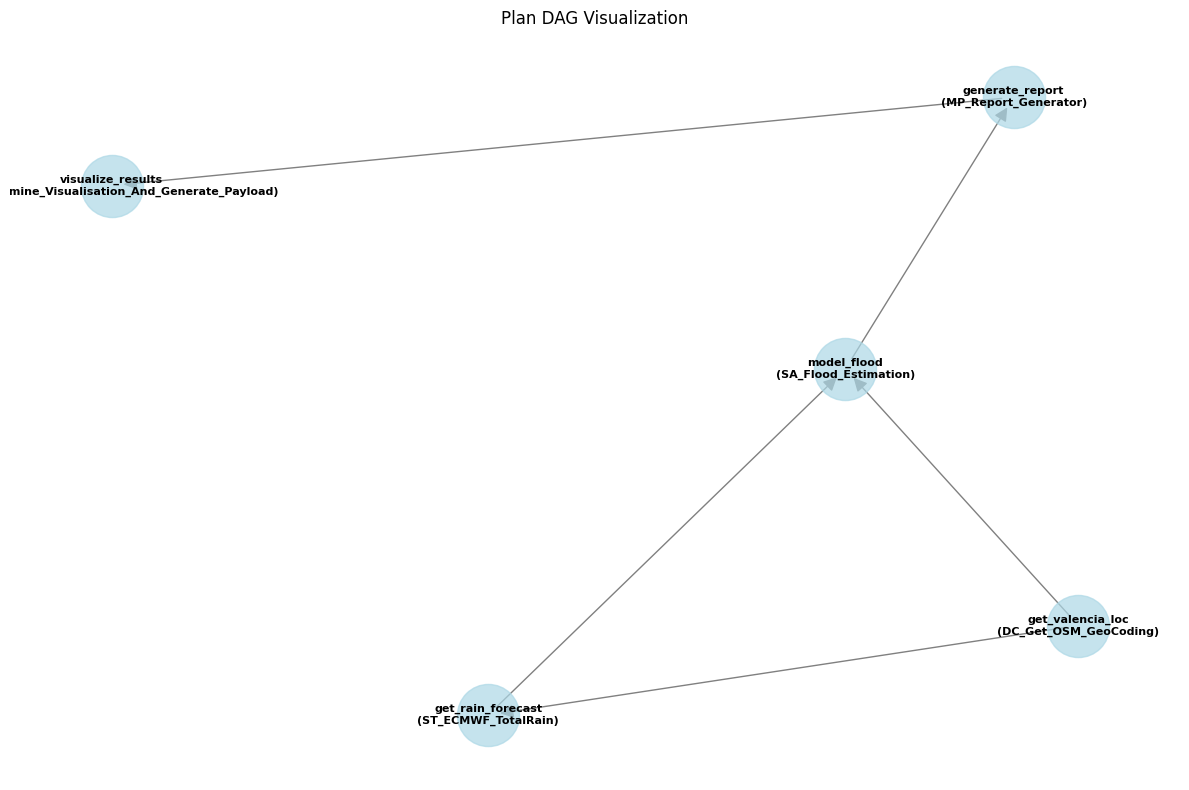}
    \caption{Internal alert reactivity flood mapping -- Automated flood risk map generation triggered by satellite imagery availability, showing the complete workflow from geographical data retrieval through precipitation forecasting to comprehensive risk visualisation.}
    \label{fig:case_study_10}
\end{figure}


\end{document}

%% file: tikz/embeddings.tex
\begin{tikzpicture}[
    box/.style={rectangle, rounded corners=5pt, draw=darkgray, thick, fill=#1, text=white, align=center, font=\small},
    stagebox/.style={rectangle, rounded corners=10pt, draw=darkgray, thick, fill=#1, minimum width=16cm, minimum height=4.5cm},
    arrow/.style={->, thick, color=arrowcolor},
    tile/.style={rectangle, draw=darkgray, fill=medgray, minimum size=0.4cm},
]
\useasboundingbox (0,-1) rectangle (16,13.5);

\node[stagebox=lightgray, anchor=south west] (stage1) at (0,9) {};
\node[anchor=north west, color=textdark, font=\bfseries] at (stage1.north west) {\hspace{0.3cm}Stage 1: Global Satellite Image Processing};

\node[ellipse, draw=textdark, thick, fill=darkgray, text=white, minimum width=2.8cm, minimum height=1.6cm] (earth) at (2.5,10.5) {Earth};

\node[tile] (t1) at (5.5,10.8) {};
\node[tile] (t2) at (6.1,10.8) {};
\node[tile] (t3) at (6.7,10.8) {};
\node[tile] (t4) at (5.5,10.2) {};
\node[tile] (t5) at (6.1,10.2) {};
\node[tile] (t6) at (6.7,10.2) {};
\node[color=textdark] at (6.1,11.4) {Satellite Tiles};

\node[box=darkgray, minimum width=2.8cm, minimum height=1.6cm] (encoder) at (9.5,10.5) {Image Encoder\\(CLAY)\\Millions of images};

\node[box=darkgray, minimum width=3.2cm, minimum height=2cm] (imgdb) at (13.5,10.5) {Image Embeddings\\Database\\(High-dimensional\\visual features)};

\draw[arrow] (earth.east) -- ++(1.5,0);
\draw[arrow] (7,10.5) -- (encoder.west);
\draw[arrow] (encoder.east) -- (imgdb.west);

\node[stagebox=lightgray, anchor=south west] (stage2) at (0,4) {};
\node[anchor=north west, color=textdark, font=\bfseries] at (stage2.north west) {\hspace{0.3cm}Stage 2: Text Description Generation (100k subset)};

\node[box=darkgray, minimum width=2.2cm, minimum height=1.6cm] (subset) at (2.5,6.5) {100k Images\\(Subset selected)};

\node[box=medgray, minimum width=2.8cm, minimum height=1.6cm] (llm) at (6.5,6.5) {LLM Description\\``Sandy dunes, arid...''};

\node[box=medgray, minimum width=2.4cm, minimum height=1.6cm] (textenc) at (10,6.5) {Text Encoder\\(Gemini)};

\node[box=darkgray, minimum width=2.6cm, minimum height=1.6cm] (textdb) at (13.5,6.5) {Text Embeddings\\(Semantic space)};

\draw[arrow] (subset.east) -- (llm.west);
\draw[arrow] (llm.east) -- (textenc.west);
\draw[arrow] (textenc.east) -- (textdb.west);

\draw[arrow, dashed, color=gray] (textdb.north) -- (imgdb.south);
\node[color=gray, font=\small] at (14.5,8) {Links};

\node[stagebox=lightgray, anchor=south west] (stage3) at (0,-1) {};
\node[anchor=north west, color=textdark, font=\bfseries] at (stage3.north west) {\hspace{0.3cm}Stage 3: Two-Stage Search Process};

\node[box=medgray, minimum width=2.8cm, minimum height=1.4cm] (query) at (2.5,2.0) {User Query\\``Show me arable\\land droughts''};

\node[box=darkgray, minimum width=3.2cm, minimum height=1.4cm] (search1) at (6.8,2.0) {Search 1: Text Space\\in Proxy Dataset};

\node[box=medgray, minimum width=2.6cm, minimum height=1.4cm] (candidates) at (10.8,2.0) {Candidate\\Images Found};

\node[box=darkgray, minimum width=3.2cm, minimum height=1.4cm] (search2) at (14.2,2.0) {Search 2: Image Space\\on Full Earth};

\node[box=textdark, minimum width=5cm, minimum height=1.4cm] (results) at (13.3,-0.0) {Similar Arable Land Drought Images};

\draw[arrow] (query.east) -- (search1.west);
\draw[arrow] (search1.east) -- (candidates.west);
\draw[arrow] (candidates.east) -- (search2.west);
\draw[arrow] (search2.south) -- (results.north);

\end{tikzpicture}

%% file: tikz/query_processing_pipeline.tex
\begin{tikzpicture}[
    node distance=1.2cm,
    box/.style={rectangle, draw, rounded corners, fill=blue!10, 
                text width=5cm, minimum height=0.8cm, align=center,
                font=\small},
    sidebox/.style={rectangle, draw, rounded corners, fill=green!10, 
                text width=3.5cm, minimum height=0.8cm, align=center,
                font=\small},
    optional/.style={rectangle, draw, dashed, rounded corners, fill=gray!10, 
                     text width=5cm, minimum height=0.8cm, align=center,
                     font=\small},
    arrow/.style={thick, -Stealth},
    ]
    
    \node[optional] (step0) {Audio transcription\\to textual query};
    
    \node[box, below=0.8cm of step0] (step1) {Natural-language query input};
    
    \node[box, below=0.8cm of step1] (step2) {Context-aware query optimisation};
    \node[sidebox, left=1cm of step2] (step2_1) {Crisis summary/RAG};
    \node[sidebox, right=1cm of step2] (step2_2) {Conversation history};
    
    \node[box, below=0.8cm of step2] (step3) {Graph planner\\with tool integration};
    
    \node[sidebox, left=1cm of step3] (tools) {Available tools description};
    \draw[arrow] (tools) -- (step3);
    
    \node[box, below=0.8cm of step3, fill=orange!10] (step4) {Graph Execution};
    
    \node[sidebox, right=1cm of step4] (datasrc) {``Digital Twin'' \&\\additional data sources};
    \draw[arrow] (datasrc) -- (step4);
    
    \node[box, below=0.8cm of step4] (step5) {Output validation by discriminator agent};
    
    \node[box, below=0.8cm of step5] (step6) {AAG presentation to expert user};
    \node[sidebox, below=0.8cm of step5, left=1cm of step5,  fill=blue!10] (step7) {Summary of result and text transmission to user};
    \node[optional, right=1cm of step5, text width=3.5cm] (step8) {Audio generation \& transmission};
    
    \draw[arrow] (step0) -- (step1);
    \draw[arrow] (step1) -- (step2);
    \draw[arrow] (step2) -- (step3);
    \draw[arrow] (step3) -- (step4);
    \draw[arrow] (step4) -- (step5);
    \draw[arrow] (step5) -- (step6);
    \draw[arrow] (step5) -- (step7);
    \draw[arrow] (step5) -- (step8);
    \draw[arrow] (step2_1) -- (step2);
    \draw[arrow] (step2_2) -- (step2);
    
    \begin{scope}[on background layer]
        \node[fit=(step0)(step6)(tools)(datasrc), draw, rounded corners, inner sep=0.3cm, fill=blue!3] {};
    \end{scope}
    
    
\end{tikzpicture}

%% file: tikz/example_AAG.tex
\begin{tikzpicture}[node distance=2cm]
    \tikzstyle{process} = [rectangle, rounded corners, minimum width=3cm, minimum height=1cm, text centered, draw=black, fill=purple!30]
    \tikzstyle{model} = [rectangle, rounded corners, minimum width=3cm, minimum height=1cm, text centered, draw=black, fill=brown!60]
    \tikzstyle{arrow} = [thick,->,>=stealth]
    \node (user) [text centered, rounded corners, draw=black, fill=white, text width=6cm] {User: "Using current weather forecasts, show me the flooding extent in Brisbane, Australia"};
    \node (geocoding) [process, below of=user] {\texttt{DC\_Get\_OSM\_GeoCoding}};
    \node (precipitation) [process, right of=geocoding, xshift=4cm] {\texttt{DC\_Get\_ECMWF\_TotalRain}};
    \node (pressure) [process, below of=precipitation] {\texttt{DC\_Get\_ECMWF\_RainPressure}};
    \node (flooding) [model, right of=precipitation, xshift=4cm] {\texttt{SA\_Flood\_Model}};
    
    \draw [arrow] (user) -- (geocoding);
    \draw [arrow] (geocoding) -- node[anchor=south, yshift=1.2em] {Brisbane Bounding Box} (precipitation);
    \draw [arrow] (geocoding) |- (pressure);
    \draw [arrow] (precipitation) -- node[anchor=south, yshift=1.2em] {Model Inputs} (flooding);
    \draw [arrow] (pressure) -| (flooding);
\end{tikzpicture}

%% file: tables/tools_by_group.tex
\newcommand{\tab}{\hspace{1cm}}

\rowcolors{2}{gray!10}{white}

\newcolumntype{L}[1]{>{}p{#1}}



\subsection{Geospatial Tools}
\begin{table}[h!]
\caption{Tools for retrieving information from OpenStreetMap (OSM). POI stands for point of interest.}
\begin{tabular}{p{6.75cm}p{6.5cm}}
\toprule
\textbf{Tool Name} & \textbf{Description} \\
\midrule
\texttt{DC\_Get\_OSM\_GeoCoding} & Convert place name to geographic data (polygon, bbox, coords) \\
\texttt{DC\_Get\_OSM\_Geospatial\_Features} & Get POIs from OSM within a polygon \\
\texttt{DC\_Get\_OSM\_POIs\_Tags\_List\_From\_Query} & Generate OSM tags from POI search query \\
\texttt{DC\_Get\_OSM\_Road\_Network\_By\_Rectangle} & Get road network within polygon \\
\bottomrule
\end{tabular}
\end{table}

\rowcolors{2}{gray!10}{white} 

\begin{table}[h!]
\caption{Tools for handling geospatial objects e.g. GeoDataFrames (GDF).}
\begin{tabular}{p{6.75cm}p{6.5cm}}
\toprule
\textbf{Tool Name} & \textbf{Description} \\
\midrule
\texttt{SA\_Intersect} & Find intersection of two GeoDataFrames \\
\texttt{SA\_Erase} & Remove portions of GDF A within GDF B boundaries \\
\texttt{SA\_Clip} & Trim GDF A to portions within GDF B boundaries \\
\texttt{SA\_Buffer} & Create buffer zone around geographic features \\
\bottomrule
\end{tabular}
\end{table}

\subsection{Alert Handling Tools}
\rowcolors{2}{gray!10}{white} 

\begin{table}[h!]
\caption{Tools for handling alerts. For example weather alerts from the MeteoAlarm API (METEO), see \url{https://www.meteoalarm.org/en/live/} }
\begin{tabular}{p{6.75cm}p{6.5cm}}
\toprule
\textbf{Tool Name} & \textbf{Description} \\
\midrule
\texttt{DC\_Collect\_METEO\_Alerts} & Collect active meteorological alerts for all regions \\
\texttt{DC\_Check\_METEO\_Alerts} & Get active meteo alerts intersecting with bbox \\
\texttt{DC\_SetMonitor\_METEO\_Alerts} & Monitor METEO alerts for region \\
\texttt{ST\_Determine\_User\_Means} & Determine which alert user refers to \\
\texttt{SA\_Check\_Inside\_Interest\_Region} & Check if alert affects user area \\
\texttt{QC\_Trigger\_Flood\_Risk\_Alert} & Determine if flood alert should trigger \\
\texttt{QC\_Test\_For\_Change} & Monitor hazard situation changes \\
\texttt{QC\_Categorize\_Change} & Define significant hazard changes \\
\texttt{MP\_Issue\_Alert} & Issue alert to \ECHO{} Alerts Board \\
\bottomrule
\end{tabular}
\end{table}

\subsection{Climate Tools}
\rowcolors{2}{gray!10}{white} 

\begin{table}[h!]
\caption{Tools for acquiring and aggregating climate information e.g. total rainfall. ECMWF is the European Centre for Medium-Range Weather Forecasts, see \url{https://www.ecmwf.int/}}
\begin{tabular}{p{6.75cm}p{6.5cm}}
\toprule
\textbf{Tool Name} & \textbf{Description} \\
\midrule
\texttt{DC\_ECMWF\_TotalRain} & Get cumulative precipitation forecast (1-10 days) \\
\texttt{DC\_ECMWF\_RasterRainForecast} & Get rainfall forecast raster data (ECMWF/ERA5) \\
\texttt{DC\_ECMWF\_RainPressure} & Get precipitation and pressure forecast \\
\bottomrule
\end{tabular}
\end{table}

\clearpage

\subsection{Hydrology Tools}
\rowcolors{2}{gray!10}{white} 

\begin{table}[h!]
\caption{Tools for extraction and processing of hydrological information.}
\begin{tabular}{p{6.75cm}p{6.5cm}}
\toprule
\textbf{Tool Name} & \textbf{Description} \\
\midrule
\texttt{DC\_Get\_Flood\_Basin} & Get flood basins intersecting with bbox \\
\texttt{ST\_Discharge\_Estimation} & Estimate river discharge from rainfall \\
\texttt{SA\_Extract\_HydroBasin} & Extract hydrological basin for location \\
\texttt{SA\_Distance\_City\_Rainfall\_Center} & Calculate distance to rainfall centre \\
\bottomrule
\end{tabular}
\end{table}

\subsection{Imagery Tools}
\rowcolors{2}{gray!10}{white} 

\begin{table}[h!]
\caption{Tools for processing of imagery already available to \ECHO.}
\begin{tabular}{p{6.75cm}p{6.5cm}}
\toprule
\textbf{Tool Name} & \textbf{Description} \\
\midrule
\texttt{DM\_Download\_Images} & Download and process satellite images from GCP \\
\texttt{DC\_Simple\_TextSearch\_for\_Images} & Find imagery by text description (no geo constraints) \\
\texttt{DC\_ImgSearch\_for\_Images} & Find imagery similar to reference image \\
\texttt{DC\_Geo\_Text\_search\_for\_Images} & Find imagery by text within geographic area \\
\bottomrule
\end{tabular}
\end{table}

\rowcolors{2}{gray!10}{white} 

\begin{table}[h!]
\caption{Tools for acquisition of satellite imagery.}
\begin{tabular}{p{6.75cm}p{6.5cm}}
\toprule
\textbf{Tool Name} & \textbf{Description} \\
\midrule
\texttt{ST\_Integrated\_Orbit\_Availability\_Estimator} & Determine best satellite imagery windows \\
\texttt{GT\_Extraction\_Scheduler} & Schedule satellite imagery extraction \\
\texttt{DC\_Integrated\_Orbit\_Extractor} & Extract satellite imagery for area/time \\
\texttt{DC\_Integrated\_Orbit\_Availability\_Retriever} & Retrieve satellite orbit planning data \\
\bottomrule
\end{tabular}
\end{table}

\subsection{Disaster Detection Tools}
\rowcolors{2}{gray!10}{white} 

\begin{table}[h!]
\caption{Tools for detection of disasters (e.g. wildfires, floods, landslides).}
\begin{tabular}{p{6.75cm}p{6.5cm}}
\toprule
\textbf{Tool Name} & \textbf{Description} \\
\midrule
\texttt{SA\_Wildfire\_Detection} & Detect wildfires from satellite imagery \\
\texttt{SA\_Landslide\_Detection} & Detect landslides from satellite imagery \\
\texttt{SA\_Flood\_Estimation} & Estimate flood extent from satellite imagery \\
\texttt{MP\_Overlayed\_Results\_On\_Sat\_Imagery} & Overlay results on satellite imagery \\
\bottomrule
\end{tabular}
\end{table}

\subsection{Evacuation Planning Tools}
\rowcolors{2}{gray!10}{white} 

\begin{table}[h!]
\caption{Tools for the planning of evacuation.}
\begin{tabular}{p{6.75cm}p{6.5cm}}
\toprule
\textbf{Tool Name} & \textbf{Description} \\
\midrule
\texttt{SA\_Generate\_Candidates} & Generate alternative safe location candidates \\
\texttt{SA\_Check\_Safe\_Route} & Check for safe evacuation routes \\
\texttt{DC\_Load\_Safe\_Zones} & Load official safe zones/shelters \\
\bottomrule
\end{tabular}
\end{table}

\newpage

\subsection{Information Aggregation Tools}
\rowcolors{2}{gray!10}{white} 
\begin{table}[h!]
\caption{Tools for the aggregation of information from various sources.}
\begin{tabular}{p{6.75cm}p{6.5cm}}
\toprule
\textbf{Tool Name} & \textbf{Description} \\
\midrule
\texttt{DC\_News\_Aggregator} & Collect disaster-related news reports \\
\texttt{DC\_Gov\_Info\_Reports\_Aggregator} & Collect government disaster reports \\
\texttt{DC\_Crowd\_Sourced\_Aggregator} & Aggregate crowd-sourced disaster info \\
\bottomrule
\end{tabular}
\end{table}

\subsection{Reporting and Communication Tools}
\rowcolors{2}{gray!10}{white} 

\begin{table}[h!]
\caption{Tools to generate reports on disasters and create visualisations.}
\begin{tabular}{p{6.75cm}p{6.5cm}}
\toprule
\textbf{Tool Name} & \textbf{Description} \\
\midrule
\texttt{MP\_Report\_Generator} & Generate comprehensive disaster reports \\
\texttt{MP\_Determine\_Visualisation\_And\_Generate\_Payload} & Generate UX payloads for visualisation \\
\bottomrule
\end{tabular}
\end{table}

\subsection{Risk Assessment Tools}
\rowcolors{2}{gray!10}{white} 

\begin{table}[h!]
\caption{Tools to assess the risks posed by hazards.}
\begin{tabular}{p{6.75cm}p{6.5cm}}
\toprule
\textbf{Tool Name} & \textbf{Description} \\
\midrule
\texttt{SA\_Determine\_Hazard} & Determine which hazard model to run \\
\texttt{SA\_Check\_Hazards} & Check for hazards at specific location \\
\texttt{SA\_Check\_Flood\_Risk} & Assess flood risk for user location \\
\bottomrule
\end{tabular}
\end{table}

\subsection{Hazard Modelling and Digital Twin Tools}
\rowcolors{2}{gray!10}{white} 

\begin{table}[h!]
\caption{Tools to collect data for digital twins, and generate flood models.}
\begin{tabular}{p{6.75cm}p{6.5cm}}
\toprule
\textbf{Tool Name} & \textbf{Description} \\
\midrule
\texttt{DC\_Collect\_Data\_For\_Twinning} & Collect data for urban digital twin \\
\texttt{SA\_Flood\_Model} & Generate flood map from rain intensity and bbox \\
\bottomrule
\end{tabular}
\end{table}

\rowcolors{2}{gray!10}{white} 

%% file: refs.bib
@article{ZHANG2024103976,
title = {GeoGPT: An assistant for understanding and processing geospatial tasks},
journal = {International Journal of Applied Earth Observation and Geoinformation},
volume = {131},
pages = {103976},
year = {2024},
issn = {1569-8432},
doi = {https://doi.org/10.1016/j.jag.2024.103976},
url = {https://www.sciencedirect.com/science/article/pii/S1569843224003303},
author = {Yifan Zhang and Cheng Wei and Zhengting He and Wenhao Yu},
}

@Manual{gdal,
    title = {{GDAL/OGR} Geospatial Data Abstraction software Library},
    author = {{GDAL/OGR contributors}},
    organization = {Open Source Geospatial Foundation},
    year = {2025},
    url = {https://gdal.org},
    doi = {10.5281/zenodo.5884351},
  }

@software{kelsey_jordahl_2019_3483425,
  author       = {Kelsey Jordahl et al},
  title        = {geopandas/geopandas: v0.6.1},
  month        = oct,
  year         = 2019,
  publisher    = {Zenodo},
  version      = {v0.6.1},
  doi          = {10.5281/zenodo.3483425},
  url          = {https://doi.org/10.5281/zenodo.3483425}
}

@software{cesium,
  author       = {Bentley Systems},
  title        = {{CesiumJS}},
  url          = {https://cesium.com/}
}

@misc{clay2024foundation,
      title        = {Clay Foundation Model: An Open Source {AI} Model for Earth},
      author       = {{Clay Foundation}},
      year         = {2024},
      howpublished = {\url{https://github.com/Clay-foundation/model}},
      note         = {Version 1.5. Pretrained Vision Transformer with masked autoencoder objective on approximately 70 million globally sampled chips from Sentinel-2, Landsat, Sentinel-1 SAR, LINZ, NAIP, and MODIS},
}

@misc{geminiteam2024gemini15unlockingmultimodal,
      title={Gemini 1.5: Unlocking multimodal understanding across millions of tokens of context}, 
      author={Gemini Team Google},
      year={2024},
      eprint={2403.05530},
      archivePrefix={arXiv},
      primaryClass={cs.CL},
      url={https://arxiv.org/abs/2403.05530}, 
}

@article{boikoAutonomousChemicalResearch2023,
  title = {Autonomous Chemical Research with Large Language Models},
  author = {Boiko, Daniil A. and MacKnight, Robert and Kline, Ben and Gomes, Gabe},
  year = {2023},
  month = dec,
  journal = {Nature},
  volume = {624},
  number = {7992},
  pages = {570--578},
  publisher = {{Springer Science and Business Media LLC}},
  issn = {0028-0836, 1476-4687},
  doi = {10.1038/s41586-023-06792-0},
  urldate = {2025-08-19},
  copyright = {https://creativecommons.org/licenses/by/4.0},
  langid = {english},
  url = {\url{https://www.nature.com/articles/s41586-023-06792-0}}
}

@article{DBLP:journals/corr/abs-2103-00020,
  author       = {Alec Radford and
                  Jong Wook Kim and
                  Chris Hallacy and
                  Aditya Ramesh and
                  Gabriel Goh and
                  Sandhini Agarwal and
                  Girish Sastry and
                  Amanda Askell and
                  Pamela Mishkin and
                  Jack Clark and
                  Gretchen Krueger and
                  Ilya Sutskever},
  title        = {Learning Transferable Visual Models From Natural Language Supervision},
  journal      = {CoRR},
  volume       = {abs/2103.00020},
  year         = {2021},
  url          = {https://arxiv.org/abs/2103.00020},
  eprinttype    = {arXiv},
  eprint       = {2103.00020},
  bibsource    = {dblp computer science bibliography, https://dblp.org}
}

@article{ghafarollahiProtAgentsProteinDiscovery2024,
  title = {{{ProtAgents}}: Protein Discovery {\emph{via}} Large Language Model Multi-Agent Collaborations Combining Physics and Machine Learning},
  shorttitle = {{{ProtAgents}}},
  author = {Ghafarollahi, Alireza and Buehler, Markus J.},
  year = {2024},
  journal = {Digital Discovery},
  volume = {3},
  number = {7},
  pages = {1389--1409},
  issn = {2635-098X},
  doi = {10.1039/D4DD00013G},
  urldate = {2025-08-19},
  langid = {english},
  url = {\url{https://xlink.rsc.org/?DOI=D4DD00013G}}
}

@book{bom1974brisbane,
  author    = {{Bureau of Meteorology}},
  title     = {Brisbane Floods January 1974: Report by Director of Meteorology},
  year      = {1974},
  publisher = {Australian Government Publishing Service},
  address   = {Canberra},
  institution = {Department of Science, Bureau of Meteorology}
}

@misc{bran2023chemcrowaugmentinglargelanguagemodels,
      title={ChemCrow: Augmenting large-language models with chemistry tools}, 
      author={Andres M Bran and Sam Cox and Oliver Schilter and Carlo Baldassari and Andrew D White and Philippe Schwaller},
      year={2023},
      eprint={2304.05376},
      archivePrefix={arXiv},
      primaryClass={physics.chem-ph},
      url={https://arxiv.org/abs/2304.05376}, 
}

@misc{szwarcman2025prithvieo20versatilemultitemporalfoundation,
      title={Prithvi-EO-2.0: A Versatile Multi-Temporal Foundation Model for Earth Observation Applications}, 
      author={Daniela Szwarcman and Sujit Roy and Paolo Fraccaro and Þorsteinn Elí Gíslason and Benedikt Blumenstiel and Rinki Ghosal and Pedro Henrique de Oliveira and Joao Lucas de Sousa Almeida and Rocco Sedona and Yanghui Kang and Srija Chakraborty and Sizhe Wang and Carlos Gomes and Ankur Kumar and Myscon Truong and Denys Godwin and Hyunho Lee and Chia-Yu Hsu and Ata Akbari Asanjan and Besart Mujeci and Disha Shidham and Trevor Keenan and Paulo Arevalo and Wenwen Li and Hamed Alemohammad and Pontus Olofsson and Christopher Hain and Robert Kennedy and Bianca Zadrozny and David Bell and Gabriele Cavallaro and Campbell Watson and Manil Maskey and Rahul Ramachandran and Juan Bernabe Moreno},
      year={2025},
      eprint={2412.02732},
      archivePrefix={arXiv},
      primaryClass={cs.CV},
      url={https://arxiv.org/abs/2412.02732}, 
}

@misc{liu2024remoteclipvisionlanguagefoundation,
      title={RemoteCLIP: A Vision Language Foundation Model for Remote Sensing}, 
      author={Fan Liu and Delong Chen and Zhangqingyun Guan and Xiaocong Zhou and Jiale Zhu and Qiaolin Ye and Liyong Fu and Jun Zhou},
      year={2024},
      eprint={2306.11029},
      archivePrefix={arXiv},
      primaryClass={cs.CV},
      url={https://arxiv.org/abs/2306.11029}, 
}

@article{Zhang_2024,
   title={RS5M and GeoRSCLIP: A Large-Scale Vision- Language Dataset and a Large Vision-Language Model for Remote Sensing},
   volume={62},
   ISSN={1558-0644},
   url={http://dx.doi.org/10.1109/TGRS.2024.3449154},
   DOI={10.1109/tgrs.2024.3449154},
   journal={IEEE Transactions on Geoscience and Remote Sensing},
   publisher={Institute of Electrical and Electronics Engineers (IEEE)},
   author={Zhang, Zilun and Zhao, Tiancheng and Guo, Yulong and Yin, Jianwei},
   year={2024},
   pages={1–23}
}

@misc{cong2023satmaepretrainingtransformerstemporal,
      title={SatMAE: Pre-training Transformers for Temporal and Multi-Spectral Satellite Imagery}, 
      author={Yezhen Cong and Samar Khanna and Chenlin Meng and Patrick Liu and Erik Rozi and Yutong He and Marshall Burke and David B. Lobell and Stefano Ermon},
      year={2023},
      eprint={2207.08051},
      archivePrefix={arXiv},
      primaryClass={cs.CV},
      url={https://arxiv.org/abs/2207.08051}, 
}

@inproceedings{pryzant2023automaticpromptoptimisationgradient,
    title = "Automatic Prompt Optimization with ``Gradient Descent'' and Beam Search",
    author = "Pryzant, Reid  and
      Iter, Dan  and
      Li, Jerry  and
      Lee, Yin  and
      Zhu, Chenguang  and
      Zeng, Michael",
    editor = "Bouamor, Houda  and
      Pino, Juan  and
      Bali, Kalika",
    booktitle = "Proceedings of the 2023 Conference on Empirical Methods in Natural Language Processing",
    month = dec,
    year = "2023",
    address = "Singapore",
    publisher = "Association for Computational Linguistics",
    url = "https://aclanthology.org/2023.emnlp-main.494/",
    doi = "10.18653/v1/2023.emnlp-main.494",
    pages = "7957--7968"
}

@inproceedings{jia2021scaling,
  title     = {Scaling Up Visual and Vision-Language Representation Learning With Noisy Text Supervision},
  author    = {Jia, Chao and Yang, Yinfei and Xia, Ye and Chen, Yi-Ting and Parekh, Zarana and Pham, Hieu and Le, Quoc V. and Sung, Yunhsuan and Li, Zhen and Duerig, Tom},
  booktitle = {Proceedings of the 38th International Conference on Machine Learning (ICML)},
  series    = {Proceedings of Machine Learning Research},
  volume    = {139},
  pages     = {4904--4916},
  year      = {2021},
  publisher = {PMLR}
}

@inproceedings{li2022blip,
  title     = {{BLIP}: Bootstrapping Language-Image Pre-training for Unified Vision-Language Understanding and Generation},
  author    = {Li, Junnan and Li, Dongxu and Xiong, Caiming and Hoi, Steven},
  booktitle = {Proceedings of the 39th International Conference on Machine Learning (ICML)},
  series    = {Proceedings of Machine Learning Research},
  volume    = {162},
  pages     = {12888--12900},
  year      = {2022},
  publisher = {PMLR}
}

@inproceedings{zhai2023sigmoid,
  title     = {Sigmoid Loss for Language Image Pre-Training},
  author    = {Zhai, Xiaohua and Mustafa, Basil and Kolesnikov, Alexander and Beyer, Lucas},
  booktitle = {Proceedings of the IEEE/CVF International Conference on Computer Vision (ICCV)},
  pages     = {11975--11986},
  year      = {2023},
  doi       = {10.1109/ICCV51070.2023.01100}
}

@inproceedings{reed2023scalemae,
  title     = {Scale-{MAE}: A Scale-Aware Masked Autoencoder for Multiscale Geospatial Representation Learning},
  author    = {Reed, Colorado J. and Gupta, Ritwik and Li, Shufan and Brockman, Sarah and Funk, Christopher and Clipp, Brian and Keutzer, Kurt and Candido, Salvatore and Uyttendaele, Matt and Darrell, Trevor},
  booktitle = {Proceedings of the IEEE/CVF International Conference on Computer Vision (ICCV)},
  pages     = {4088--4099},
  year      = {2023},
  doi       = {10.1109/ICCV51070.2023.00378}
}

@misc{xiong2024dofa,
  title         = {Neural Plasticity-Inspired Multimodal Foundation Model for Earth Observation},
  author        = {Xiong, Zhitong and Wang, Yi and Zhang, Fahong and Stewart, Adam J. and Hanna, Jo{\"e}lle and Borth, Damian and Papoutsis, Ioannis and Le Saux, Bertrand and Camps-Valls, Gustau and Zhu, Xiao Xiang},
  year          = {2024},
  eprint        = {2403.15356},
  archivePrefix = {arXiv},
  primaryClass  = {cs.CV},
  doi           = {10.48550/arXiv.2403.15356}
}

@misc{xiong2025dofaclip,
  title         = {{DOFA-CLIP}: Multimodal Vision-Language Foundation Models for Earth Observation},
  author        = {Xiong, Zhitong and Wang, Yi and Yu, Weikang and Stewart, Adam J. and Zhao, Jie and Lehmann, Nils and Dujardin, Thomas and Yuan, Zhenghang and Ghamisi, Pedram and Zhu, Xiao Xiang},
  year          = {2025},
  eprint        = {2503.06312},
  archivePrefix = {arXiv},
  primaryClass  = {cs.CV},
  doi           = {10.48550/arXiv.2503.06312}
}

@inproceedings{wang2024skyscript,
  title     = {{SkyScript}: A Large and Semantically Diverse Vision-Language Dataset for Remote Sensing},
  author    = {Wang, Zhecheng and Prabha, Rajanie and Huang, Tianyuan and Wu, Jiajun and Rajagopal, Ram},
  booktitle = {Proceedings of the AAAI Conference on Artificial Intelligence},
  volume    = {38},
  number    = {6},
  pages     = {5805--5813},
  year      = {2024},
  doi       = {10.1609/aaai.v38i6.28393}
}

@inproceedings{bastani2023satlaspretrain,
  title     = {{SatlasPretrain}: A Large-Scale Dataset for Remote Sensing Image Understanding},
  author    = {Bastani, Favyen and Wolters, Piper and Gupta, Ritwik and Ferdinando, Joe and Kembhavi, Aniruddha},
  booktitle = {Proceedings of the IEEE/CVF International Conference on Computer Vision (ICCV)},
  pages     = {16772--16782},
  year      = {2023}
}

@article{hong2024spectralgpt,
  title   = {{SpectralGPT}: Spectral Remote Sensing Foundation Model},
  author  = {Hong, Danfeng and Zhang, Bing and Li, Xuyang and Li, Yuxuan and Li, Chenyu and Yao, Jing and Yokoya, Naoto and Li, Hao and Ghamisi, Pedram and Jia, Xiuping and Plaza, Antonio and Gamba, Paolo and Benediktsson, Jon Atli and Chanussot, Jocelyn},
  journal = {IEEE Transactions on Pattern Analysis and Machine Intelligence},
  volume  = {46},
  number  = {8},
  pages   = {5227--5244},
  year    = {2024},
  doi     = {10.1109/TPAMI.2024.3362475}
}

@article{xiao2024foundation,
  title   = {Foundation Models for Remote Sensing and Earth Observation: A Survey},
  author  = {Xiao, Aoran and Xuan, Weihao and Wang, Junjue and Huang, Jiaxing and Tao, Dacheng and Lu, Shijian and Yokoya, Naoto},
  journal = {IEEE Geoscience and Remote Sensing Magazine},
  year    = {2025},
  note    = {In press},
  doi     = {10.1109/MGRS.2025.3590471}
}

@inproceedings{yao2023react,
  title     = {{ReAct}: Synergizing Reasoning and Acting in Language Models},
  author    = {Yao, Shunyu and Zhao, Jeffrey and Yu, Dian and Du, Nan and Shafran, Izhak and Narasimhan, Karthik and Cao, Yuan},
  booktitle = {Proceedings of the 11th International Conference on Learning Representations (ICLR)},
  year      = {2023}
}

@inproceedings{schick2023toolformer,
  title     = {Toolformer: Language Models Can Teach Themselves to Use Tools},
  author    = {Schick, Timo and Dwivedi-Yu, Jane and Dess{\`i}, Roberto and Raileanu, Roberta and Lomeli, Maria and Zettlemoyer, Luke and Cancedda, Nicola and Scialom, Thomas},
  booktitle = {Advances in Neural Information Processing Systems 36 (NeurIPS 2023)},
  year      = {2023}
}

@inproceedings{wu2024autogen,
  title     = {{AutoGen}: Enabling Next-Gen {LLM} Applications via Multi-Agent Conversation},
  author    = {Wu, Qingyun and Bansal, Gagan and Zhang, Jieyu and Wu, Yiran and Li, Beibin and Zhu, Erkang and Jiang, Li and Zhang, Xiaoyun and Zhang, Shaokun and Liu, Jiale and Awadallah, Ahmed Hassan and White, Ryen W. and Burger, Doug and Wang, Chi},
  booktitle = {Proceedings of the 1st Conference on Language Modeling (COLM)},
  year      = {2024}
}

@inproceedings{bonafilia2020sen1floods11,
  title     = {{Sen1Floods11}: A Georeferenced Dataset to Train and Test Deep Learning Flood Algorithms for {Sentinel-1}},
  author    = {Bonafilia, Derrick and Tellman, Beth and Anderson, Tyler and Issenberg, Erica},
  booktitle = {Proceedings of the IEEE/CVF Conference on Computer Vision and Pattern Recognition (CVPR) Workshops},
  pages     = {210--211},
  year      = {2020}
}

@misc{gupta2019xbd,
  title         = {{xBD}: A Dataset for Assessing Building Damage from Satellite Imagery},
  author        = {Gupta, Ritwik and Hosfelt, Richard and Sajeev, Sandra and Patel, Nirav and Goodman, Bryce and Doshi, Jigar and Heim, Eric and Choset, Howie and Gaston, Matthew},
  year          = {2019},
  eprint        = {1911.09296},
  archivePrefix = {arXiv},
  primaryClass  = {cs.CV},
  doi           = {10.48550/arXiv.1911.09296}
}

@article{mateogarcia2021worldfloods,
  title   = {Towards global flood mapping onboard low cost satellites with machine learning},
  author  = {Mateo-Garcia, Gonzalo and Veitch-Michaelis, Joshua and Smith, Lewis and Oprea, Silviu Vlad and Schumann, Guy and Gal, Yarin and Baydin, At{\i}l{\i}m G{\"u}ne{\c{s}} and Backes, Dietmar},
  journal = {Scientific Reports},
  volume  = {11},
  number  = {1},
  pages   = {7249},
  year    = {2021},
  doi     = {10.1038/s41598-021-86650-z}
}

@inproceedings{karpukhin2020dpr,
  title     = {Dense Passage Retrieval for Open-Domain Question Answering},
  author    = {Karpukhin, Vladimir and O{\u{g}}uz, Barlas and Min, Sewon and Lewis, Patrick and Wu, Ledell and Edunov, Sergey and Chen, Danqi and Yih, Wen-tau},
  booktitle = {Proceedings of the 2020 Conference on Empirical Methods in Natural Language Processing (EMNLP)},
  pages     = {6769--6781},
  year      = {2020},
  publisher = {Association for Computational Linguistics},
  doi       = {10.18653/v1/2020.emnlp-main.550}
}

@inproceedings{khattab2020colbert,
  title     = {{ColBERT}: Efficient and Effective Passage Search via Contextualized Late Interaction over {BERT}},
  author    = {Khattab, Omar and Zaharia, Matei},
  booktitle = {Proceedings of the 43rd International ACM SIGIR Conference on Research and Development in Information Retrieval (SIGIR)},
  pages     = {39--48},
  year      = {2020},
  doi       = {10.1145/3397271.3401075}
}

@inproceedings{khattab2024dspy,
  title     = {{DSPy}: Compiling Declarative Language Model Calls into State-of-the-Art Pipelines},
  author    = {Khattab, Omar and Singhvi, Arnav and Maheshwari, Paridhi and Zhang, Zhiyuan and Santhanam, Keshav and Vardhamanan, Sri and Haq, Saiful and Sharma, Ashutosh and Joshi, Thomas T. and Moazam, Hanna and Miller, Heather and Zaharia, Matei and Potts, Christopher},
  booktitle = {Proceedings of the 12th International Conference on Learning Representations (ICLR)},
  year      = {2024}
}

@inproceedings{yang2024opro,
  title     = {Large Language Models as Optimizers},
  author    = {Yang, Chengrun and Wang, Xuezhi and Lu, Yifeng and Liu, Hanxiao and Le, Quoc V. and Zhou, Denny and Chen, Xinyun},
  booktitle = {Proceedings of the 12th International Conference on Learning Representations (ICLR)},
  year      = {2024}
}

@inproceedings{weyand2016planet,
  title     = {{PlaNet} - Photo Geolocation with Convolutional Neural Networks},
  author    = {Weyand, Tobias and Kostrikov, Ilya and Philbin, James},
  booktitle = {Computer Vision -- ECCV 2016},
  series    = {Lecture Notes in Computer Science},
  volume    = {9912},
  pages     = {37--55},
  year      = {2016},
  publisher = {Springer},
  doi       = {10.1007/978-3-319-46484-8_3}
}

@inproceedings{lacoste2023geobench,
  title     = {{GEO-Bench}: Toward Foundation Models for Earth Monitoring},
  author    = {Lacoste, Alexandre and Lehmann, Nils and Rodriguez, Pau and Sherwin, Evan David and Kerner, Hannah and L{\"u}tjens, Bj{\"o}rn and Irvin, Jeremy Andrew and Dao, David and Alemohammad, Hamed and Drouin, Alexandre and Gunturkun, Mehmet and Huang, Gabriel and Vazquez, David and Newman, Dava and Bengio, Yoshua and Ermon, Stefano and Zhu, Xiao Xiang},
  booktitle = {Advances in Neural Information Processing Systems 36 (NeurIPS 2023) Datasets and Benchmarks Track},
  year      = {2023}
}
